\documentclass[a4paper,10pt]{article}

\usepackage[british]{babel}

\usepackage[utf8]{inputenc}
\usepackage[T1]{fontenc}

\usepackage{libertine}
\usepackage{inconsolata}

\usepackage[
   protrusion=true
  ,expansion
  ,babel=true]{microtype}
\usepackage[autostyle=true]{csquotes}
\usepackage{expex}
\usepackage{graphicx}
\usepackage{longtable}
\usepackage{wrapfig}
\usepackage{rotating}
\usepackage[normalem]{ulem}
\usepackage{amsmath}
\usepackage{amssymb}
\usepackage{capt-of}
\usepackage[hidelinks]{hyperref}
\usepackage{orcidlink}
\usepackage{booktabs}
\usepackage{tabularx}

\usepackage{tikz}
\usetikzlibrary{shapes,arrows,positioning,calc,matrix,backgrounds,graphs,quotes}
\tikzset{align at top/.style={baseline=(current bounding box.north)}}
\tikzset{align at mid/.style={baseline=(current bounding box.east)}}

\usepackage{listings}
\usepackage[%
   style=authoryear
  ,dashed=false
  ,maxbibnames=100
]{biblatex}
\title{Conversational Implicatures: Modelling Relevance Theory Probabilistically}

\author{%
  Christoph Unger \orcidlink{0000-0002-3187-8719}
\and 
  Hendrik Buschmeier \orcidlink{0000-0002-9613-5713}}

\date{%
  Faculty of Linguistics and Literary Studies,\\
  Bielefeld University, Bielefeld, Germany
  \texttt{\{christoph.unger|hbuschme\}@uni-bielefeld.de}
}

\begin{document}

\maketitle
\gathertags

\begin{abstract}
  Recent advances in Bayesian probability theory and its application to cognitive science in combination with the development of a new generation of computational tools and methods for probabilistic computation have led to a `probabilistic turn' in pragmatics and semantics. In particular, the framework of Rational Speech Act theory has been developed to model broadly Gricean accounts of pragmatic phenomena in Bayesian terms, starting with fairly simple reference games and covering ever more complex communicative exchanges such as verbal syllogistic reasoning. This paper explores in which way a similar Bayesian approach might be applied to relevance-theoretic pragmatics (Sperber \& Wilson, 1995) by study a paradigmatic pragmatic phenomenon: the communication of implicit meaning by ways of (conversational) implicatures.
\end{abstract}

\section{Introduction}
\label{sec:introduction}

Recent advances in Bayesian probability theory and its application to cognitive science \parencite[e.g.,][]{tenenbaumTheorybasedBayesianModels2006,tenenbaumHowGrowMind2011} in combination with the development of a new generation of computational tools and methods for probabilistic computation \parencite{mansinghkaNativelyProbabilisticComputation2009,fierensInferenceLearningProbabilistic2015} have led to what is sometimes called a `probabilistic turn' in pragmatics and semantics \parencite{erkProbabilisticTurnSemantics2022}. In particular, the framework of Rational Speech Act theory \parencite{degenRationalSpeechAct2023} has been developed to model broadly Gricean accounts of pragmatic phenomena in Bayesian terms, starting with fairly simple reference games \parencite{frankPredictingPragmaticReasoning2012} and covering ever more complex communicative exchanges such as verbal syllogistic reasoning \parencite{tesslerLogicProbabilityPragmatics2022}.

In this paper we explore in which way a similar Bayesian approach might be applied to relevance-theoretic pragmatics \parencite{sperberRelevance1995,clarkRelevanceTheory2013}. To this end, we study the perhaps most paradigmatic pragmatic phenomenon, the communication of implicit meaning by ways of (conversational) implicatures.

Our investigation proceeds according to the following plan: in Section \Ref{sec:implicature}, we discuss the phenomenon and main properties of implicatures in informal terms. Section \Ref{sec:compr-impl} reviews the relevance theoretic account of implicature comprehension and how this is intertwined with explicature comprehension. In Section \Ref{sec:observ-cont-compr} we explore some of the depths of the theoretical concepts involved and explain how we propose to capture these notions in the probabilistic model we aim to construct. Finally, in Section \Ref{sec:devel-prob-model} we explain in detail a probablistic model of implicature comprehension implemented in the probabilistic logic programming language ProbLog.

\section{Introducing Implicature}
\label{sec:implicature}

Consider the following conversation between Peter and Mary:

\pex<impl>
  \a<implpeter>\emph{Peter: } Do you want some coffee? 

  \a<implmary>\emph{Mary: } I don't want to drink energizing drinks at this time of day.
\xe
Intuitively, what Mary intends to communicate to Peter can be paraphrased as in (\getfullref{implparaphrase}):

\pex<implparaphrase>
  Mary does not want to drink coffee at 6 pm because coffee is an energizing drink and she does not want to drink energizing drinks at 6 pm.
\xe

Let us re-order the component conjuncts of this paraphrase of Mary's intended meaning as in (\getfullref{marysplicatures}) and ask about the relationship of each component to Mary's utterance in (\getfullref{impl.implmary}):

\pex<marysplicatures>
  \a<implprem> Coffee is an energizing drink.

  \a<implconc> Mary does not want to drink coffee at this time of day.

  \a<explicature> Mary does not want to drink energizing drinks at 6 pm. 
\xe

(\getfullref{marysplicatures.explicature}) is derived from Mary's sentence in (\getfullref{impl.implmary}) by assigning reference to the pronoun \emph{I} and resolving the time reference of the expression \emph{at this time of day} in the context of the conversational setting. As a result, we can assign truth conditions for the resulting representation (\getfullref{marysplicatures.explicature}), something we can not do for the linguistic representation (\getfullref{impl.implmary}) (the sentence uttered by Mary). Notice, however, that (\getfullref{marysplicatures.explicature}) has a close relationship to the logical form of (\getfullref{impl.implmary}): it is derived from the logical blueprint given by the sentence through contextual enrichment and reference resolution. Let us call this process of contextually supplementing logical forms to the points where the resulting representations carries truth conditions the \emph{development} of the logical form of the utterances \parencite{sperberRelevance1995}. Representations that the audience derives as a result of such a process of the development of the logical form of an utterance are \emph{explicitly} communicated and are called \emph{explicatures} \parencite{sperberRelevance1995}.

However, (\getfullref{marysplicatures.explicature}) alone does not suffice to answer Peter's question. Yet Mary ostensibly presents her utterance of (\getfullref{impl.implmary}) as an answer to Peter's question. In order to explain how Mary could have intended her utterance to answer Peter's question, Peter has to supply the assumption (\getfullref{marysplicatures.implprem}) as a contextual premise. From this contextual premise and the respresentation (\getfullref{marysplicatures.explicature}) Peter can conclude (\getfullref{marysplicatures.implconc}) that \emph{Mary does not want to drink coffee at 6 pm}. Notice that neither (\getfullref{marysplicatures.implprem}) nor (\getfullref{marysplicatures.implconc}) are derived by further enriching the logical form of Mary's utterance; rather, (\getfullref{marysplicatures.implprem}) is an independent proposition that together with the proposition conveyed by Mary's utterance yields the conclusion (or in other words, contextual implication) (\getfullref{marysplicatures.implconc}). Implicit premises and conclusions that need to be contextually supplied in order for the audience to comprehend the speaker's meaning, but are not the result of the development of a logical form, are fully \emph{implicitly} communicated and are called \emph{implicatures.}

\section{Comprehending implicatures}
\label{sec:compr-impl}

\subsection{Verbal communication works by ostension}
\label{sec:verb-comm-works}

What is Mary achieving when she utters (\getfullref{impl.implmary})? As we have seen, she has not encoded all aspects of the meaning she wanted to convey to Peter into her utterance. Peter has to supplement the logical form he can decode from the linguistic form of Mary's utterance in two ways:

\pex<supplement>
  \a<expl> Develop the logical form of the utterance into a proposition by reference resolution of indexicals using contextual evidence. 

  \a<impl> Selecting a suitable implicit premise as a contextual assumption for deriving an implicit conclusion that answers Peter's question (and thereby allows him `to see the point' of Mary's utterance).
\xe

What Mary has done by uttering (\getfullref{impl.implmary}) is that she has provided Peter with partial evidence for inferring her intended meaning. In other words, comprehension is an exercise in intention attribution.

That comprehending verbal communication is essentially an act of intention attribution was one of the ground breaking insights of the philosopher Paul Grice, discussed in his seminal paper \emph{Meaning} \parencite{griceMeaning1957}. \textcite{sperberRelevance1995} have taken up these ideas and refined them by arguing that in verbal communication the communicator makes overt a layered set of intentions as follows:

\pex<intention>
  \a<informative>\emph{Informative intention: } to make manifest or more manifest to the audience a set of assumptions \textbf{I}. \parencite[58]{sperberRelevance1995}

  \a<communicative>\emph{Communicative intention: } to make it mutually manifest to audience and communicator that the communicator has this informative intention. \parencite[61]{sperberRelevance1995}
\xe

Central in these definitions is the concept of the \emph{manifestness} of a piece of information \parencite[38--46]{sperberRelevance1995}. We quote their definition from a later paper in which the authors elaborate further on the impact of the notion of \emph{manifestness}:

\pex<manifestness>
  A proposition\footnote{Directly after presenting this definition, Sperber and Wilson make the following comment: ``In \emph{Relevance, } we talked not of propositions but of assumptions as being manifest. Either term will do, and so would ‘pieces of information’. What we are talking about are things that can be true or false and that, when they are true, are facts.'' \parencite[134]{sperberSpeakersMeaning2015}} is manifest to an individual at a given time to the extent that he is likely to some positive degree to entertain it and accept it as true. \parencite[134]{sperberSpeakersMeaning2015}
\xe

The manifestness of a piece of information is influenced by two factors: the epistemic status of the piece of information (how likely it is to be accepted as true), and its salience or accessibility (how likely it is to be considered at all). Manifestness is the joint influence of these factors and is a measure for the probability that the given piece of information will influence an individual's believes or decisions \parencite[134]{sperberSpeakersMeaning2015}.

The communicator's informative intention was defined as the intention to increase the manifestness of a set of assumptions to an audience. This set of assumptions is the content of the speaker's meaning. In our example it is the set of assumptions listed in (\getfullref{marysplicatures})\footnote{%
  In fact, this is a simplified statement that will be revised shortly}.
In other words, Mary's informative intention is to increase the likelihood that Peter will mentally represent these assumptions as Mary's intended meaning, ideally even so much that he believes them.

The effect of Mary's communicative intention is to make it (more) manifest to Peter not only that she intends to make the assumptions in (\getfullref{marysplicatures}) (more) manifest to him, but also the fact that Mary intends Peter to recognise that she intends to do so. In other words, a communicator making her informative intention mutually manifest ensures that this intention of hers becomes overt to the audience. Overtly intentional acts are processed in a special way by organisms with minds that strife for efficiency in information processing and knowledge representation, as we shall see shortly in the following sections.

\subsection{Inputs to cognitive processes compete for relevance}
\label{sec:inputs-cogn-proc}

\textcite[261--262]{sperberRelevance1995} point out that as a biological function, cognition is under preasure to process inputs \emph{efficiently}. To measure the efficiency of input processing, relevance theory introduces the notion of \emph{relevance} as a technical term. An input is relevant to the extent that it leads to the most \emph{positive cognitive effects} for the least processing effort involved in deriving these effects. Positive cognitive effects are improvements of the mind's representation of the world, and may include confirmation (strengthening) of existing knowledge, deletion of assumptions that contradict existing knowledge but are false (contradiction and elimination), or the acquisition of new information as a result of combining the input with existing assumptions \parencite[contextual implication;][118--132]{sperberRelevance1995}.

Suppose you are sitting at your desk writing a paper. You pause and look at the wall in order to give your eyes rest from looking at the computer screen. There happen to be too small black dots at the wall. You hardly notice them and most likely don't pay attention to them. Then suddenly you have the impression that one of the dots moves around ever so slightly. You sit up and look more closely at this dot. Why? Because the input \textsc{The dot may have moved} easily gives rise to the following thought chain: if it indeed has moved, it may be a living thing, an insect, for instance a spider. And if it is a spider, you don't want it to be there, you want to remove it. In other words: the input has become more relevant to you: there are easily available cognitive effects.

At the same time, you stop paying attention to the other dot that hasn't moved. But you do get up, grap a sheet of paper and go to the wall to check. Which dot will you investigate first? Surely the dot that you think has moved. Why not the other? After all, this other dot is of roughly the same size and shape. But it hasn't given rise to more cognitive effects as the other dot has.

This example illustrates a descriptive generalisation that relevance theory makes about cognition in general: the mind tends to pay attention to inputs that are most relevant. \textcite[260]{sperberRelevance1995} put it in these terms:

\pex<cogpor>
  Human cognition tends to be geared to the maximisation of relevance.
\xe

By allocating scarce processing resources primarily to such inputs that promise to be most relevant, information is processed efficiently.
\textcite[134--135]{sperberSpeakersMeaning2015} point out that there is a close connection between the notions of manifestness and of relevance:

\begin{quote}
  However, manifestness is this ceteris paribus probability of influence, rather than the factors that contribute to it. The same point can be made on the basis of the Cognitive Principle of Relevance (`Human cognition tends to be geared to the maximisation of relevance'): Given a belief which has a cognitive effect and which is therefore relevant in a categorical sense, its relevance will be comparatively greater 1) to the extent that its processing is less costly because it is more salient, and 2) to the extent that its epistemic strength is greater. If a proposition is relevant at all, then the greater its manifestness, the greater its relevance.
\end{quote}

\subsection{Overtly intentional inputs claim a good measure of relevance}
\label{sec:overtly-intent-input}

Overtly intentional inputs are relevant in terms of the intentions that the communicator expects the audience to infer. This means essentially that once the audience has grounds to attribute a communicative intention to a communicator, it is forced to put in the effort to identify an informative intention. Given that the mind tends to allocate processing resources for those inputs that are most relevant, this means that overtly intentional inputs raise the expectation that they are relevant to a specific degree: they must at least be relevant enough for the audience to be worth the attention. But given the mind's tendency to maximise relevance, it will be beneficial for the communicator if the overtly intentional input is more relevant to the audience than this minimum requirement. On the other hand, rational audiences can not expect perfection: communicators may be limited in their \emph{abilities} and \emph{preferences}: in regard to their abilities, communicator's may not think of the best words, the easiest grammatical expression, or simply have an erroneous assumption of what is relevant for the audience. In regard to their preferences, communicators may want to conform to societal expectations, respect cultural taboos, and so on. All in all, overtly intentional inputs promise to be \emph{optimally relevant}, where \emph{optimal relevance} is a level of relevance with a lower bound of being relevant just enough to be worth the attention, and an upper bound of being as relevant as the communicator's abilities and preferences allow. This is expressed as the communicative principle of relevance:

\pex<commpor>
  Commumicative Principle of Relevance \\
  Every [act of overtly intensional communication] conveys a presumption of its own optimal relevance. \parencite[Slightly adapted from][612]{wilsonRelevanceTheory2004}
\xe

\subsection{Processing overtly intentional inputs with an efficient heuristic procedure}
\label{sec:proc-overtly-intentional}
  
Because overtly intentional inputs come with an implicit claim that they are optimally relevant to the audience, the informative intention undrlying them can be found following the heuristic procedure in (\nextx):

 \pex<relproc>
  \emph{The relevance-theory processing heuristic}
  
  \a<start> Access the most easily accessible interpretive hypothesis, consisting of a hypothesis about the implicit import, a hypothesis about the fine-tuning of explicit meaning, together with the contextual assumptions that support these hypotheses.
  
  \a<check> Check if the stimulus, under this interpretation, yields enough cognitive effects of the right kind to meet the audience's expectation of relevance.
  
  \a<caseyes> If it does meet relevance expectations, the audience is justified to assume that this is the interpretation intended by the communicator.
  
  \a<caseno> If it does not meet relevance expectations, the audience may access other interpretive hypotheses in the order of accessibility until an interpretation satisfying relevance expectations is found, or the search is abandoned because the processing effort is deemed too high.
\xe

Let us apply this processing heuristic to example (\getfullref{impl.implmary}) and see how this procedure explains how the audience arrives spontaneously and intuitively at the interpretations in (\getfullref{marysplicatures}).

\subsection{Processing implicatures}
\label{sec:proc-impl}

At the time Mary in example (\getfullref{impl.implmary}) is taking her turn in the conversation, Peter (the audience) has specific expectations about the content of cognitive effects that any utterance of Mary must have in order to be relevant to him: it must be interpretable as providing a positive or negative answer to the question Peter asked, whether Mary wants to have some coffee. Howver, the utterance itself does not say anything about coffee itself. But it does convey the concept of \textsc{energizing drinks}. This concept bundles our encyclopaedic knowledge about energizing drinks, which presumably includes a list of examples:

 \pex<energizingdrinks>
 \textsc{energizing drinks} \\
   \ldots{} \\
   Coffee is an energizing drink. \\
   Cola is an energizing drink. \\
   Redbull is an energizing drink. \\
   Tea is an energizing drink. \\
   Peppermint tea is an energizing drink. \\
   \ldots{}
 \xe

Since the concept \textsc{coffee} has been activated by the previous conversational turn (Peter's question), the assumption \textsc{Coffee is an energizing drink} is more highly activated, and thus easier accessible than, the other pieces of encyclopaedic information associated with the concept \textsc{energizing drinks}. Assuming that this idea is intended to be used as context, the utterance can be interpreted along the lines in (\getfullref{implanalysisindira}):

\pex<implanalysisindira>
  \a<context> Coffee is an energizing drink.

  \a<premisea> If coffee is an energizing drink, Mary does not want to drink coffee.
  
  \a<conclusion> Mary does not want to drink coffee. 
\xe

Accessing assumption (\getfullref{implanalysisindira.context}) and using it as context, this assumption can be used as second premise in a syllogism using the conditional (\getfullref{implanalysisindira.premisea}) to support the conclusion in (\getfullref{implanalysisindira.conclusion}). This conclusion in turn provides a cognitive effect of the expeted kind. Since these interpretive hypotheses are very easy to access, and the utterance achieves sufficient cognitive effects of the expected kind, the audience (Peter) is entitled to attribute to Mary the intention to communicate the thoughts in (\getfullref{implanalysisindira}).

However, notice that the utterance (\getfullref{impl.implmary}) has raised the accessibility of other contextual assumptions as well:

\pex<otherenergydrinks>
  \a<cola> Coke is an energizing drink.

  \a<redbull> Redbull is an energizing drink.
  
  \a<tea> Tea is an energizing drink.
  
  \a<peppermint> Peppermint tea is an energizing drink. 
\xe

Though not as highly accessible as, and hence less manifest than, the assumption \emph{Coffee is an energizing drink}, at least some of them are still very easily accessible, and by inferential processes closely parallel to the one in (\getfullref{implanalysisindira}), these in turn lead to the conclusions in (\getfullref{noother}):

\pex<noother>
  \a<noc> Mary does not want to drink coke.

  \a<nor> Mary does not want to drink redbull. 

  \a<not> Mary does not want to drink tea.

  \a<nop> Mary does not want to drink peppermint tea. 
\xe

While the interpretation in (\getfullref{implanalysisindira}) alone would suffice to make Mary's utterance relevant enough to Peter to be worth the attention, expanding the context slightly by accessing, for instance,  (\getfullref{otherenergydrinks.cola}) and (\getfullref{otherenergydrinks.redbull}) will yield more cognitive effects. Assuming that doing so does not obviously violate the audience's estimation of the communicator's abilities and preferences, the audience is licensed to include these assumptions and the inferences they support in their interpretive hypothesis, although with less confidence than the more easily accessible (and hence more relevant and more manifest) assumption \emph{B does not want to drink coffee}. What about the still less accessible (and hence less manifest) assumptions (\getfullref{noother.not}) and (\getfullref{noother.nop})? This depends on the audience's expectations of relevance, in particular on how much processing effort the audience is prepared to tolerate.

So in other words, this analysis suggest that all the possible contextual assumptions linked to the encyclopaedic entries of the concept \textsc{energizing drink}, and the contextual implications they give rise to, are becoming more manifest to Peter as a result of processing Mary's utterance following the relevance-theoretic comprehension procedure. They are becoming more manifest to different degrees, as they differ in accessibility. Recall that the manifestness of a piece of information is a measure of the probability that the piece of information will play a causal role in the individual's mental life. This probability itself does not entail that the individual actually mentally represents this piece of information; it merely entails that the individual is capable, in principle, to mentally represent this piece of information and assess its truth. Thus, the high probability that (\getfullref{marysplicatures.implprem}) and (\getfullref{marysplicatures.implconc}) will be accepted by Peter as part of Mary's intended meaning does not itself guarantee that Peter will do so; but it will give Peter a strong incentive for doing so. \textcite[199]{sperberRelevance1995} say that implicatures such as (\getfullref{marysplicatures.implprem}) and (\getfullref{marysplicatures.implconc}) are \emph{strongly communicated} and constitute \emph{strong implicatures}. Peter's incentive for representing implicatures such as (\getfullref{otherenergydrinks.peppermint}) and (\getfullref{noother.nop}) is much weaker. \textcite[199]{sperberRelevance1995} speak of \emph{weakly communicated} assumptions and call them \emph{weak implicatures}. Obviously, the distinction between strong and weak implicatures is not a categorical one; rather, there is a cline between them.

For the purposes of this paper, we will assume that Peter represents only the implicatures (\getfullref{marysplicatures.implprem}) and (\getfullref{marysplicatures.implconc}) and attributes them to Mary's informative intention, the implicatures for which Peter has the strongest grounds for attributing them to Mary. Still, the weaker implicatures have a certain likelihood to influence Peter's behaviour: realising that Mary declined his offer of a cup of coffee, he may want to offer her something else instead. At this point, the weaker communicated implicatures may influence his alternative offer.

The comprehension process of the implicatures in example (\getfullref{impl.implmary}) are summarised in table \Ref{tab:impicature-compr-coffee}.

\begin{table}
  \centering
  \begin{tabularx}{\textwidth}{XX}
    \toprule
    Representation & Comment \\ 
    \midrule
      (a) Mary has said to Peter ``I don't want to drink energizing drinks at this time of day.'' 
    & Embedding of the decoded incomplete logical form of Mary's utterance into a description of Mary's overtly intentional act. \\
      (b) Mary's utterance is optimally relevant to Peter. 
    & Expectation raised by the overt nature of Mary's intentional act and acceptance of the presumption of relevance it conveys.\\
      (c) Mary's utterance is relevant to Peter by specifying whether or not Mary wants to drink some coffee now at 6 pm. 
    & Fine-tuned expectation of relevance raised by (b) in the light of Peter's preceding question. \\
      (d) Coffee is an energizing drink. 
    & Encyclopaedic entry of the concept \textsc{energizing drink}. Accepted as an implicit premise, an implicature of Mary's utterance. \\
      (e) Mary does not want to drink energising drinks at 6 pm.
    & Proposition derived by reference resolution of the pronoun \emph{I} and fixing the time reference of the phrase \emph{at this time of day} to 6 pm, the time of the talk exchange. Accepted as the explicature of Mary's utterance. \\
      (f) Mary does not want to drink coffee at 6 pm. 
    & Implicit conclusion from (d) and (e). Satisfies expectation (c) and (b). Accepted as an implicature of Mary's utterance. \\
      (g) Coke is an energising drink. 
    & Other encyclopaedic entry of the concept \textsc{energizing drink}. Serves as another implicit premise. \\
      (h) Mary does not want to drink coke at 6 pm. 
    & Implicit conclusion from (g) and (e). Increases satisfaction of expectation (c) and (b). Accepted as a less manifest implicature of Mary's utterance. \\\relax
      [The encyclopaedic entries of the concept \textsc{energising drink} may provide other premises similar to (d) and (g)] 
    & [Potentially supporting further implicit conclusions, each less manifest the less accessible they are, and the less processing effort Peter is prepared to invest.] \\
  \bottomrule
\end{tabularx}
  \caption[comprehension summary]{Summary table of implicature comprehension in example (\getfullref{impl.implmary})}
  \label{tab:impicature-compr-coffee}
\end{table}

\section{Observing the contours of the comprehension heuristic}
\label{sec:observ-cont-compr}

\subsection{Inference under uncertainty: heuristics and probabilistic models}
\label{sec:infer-under-uncert}

\subsubsection{Communicaton and inference under uncertainty}
\label{sec:comm-infer-under}

The philosopher Grice was intrigued by the question why logical connectives in natural languages, like \emph{and} in English, seem to deviate in their meaning from the logical operator $\land$, as illustrated by contrast between (\getfullref{nland.seq}) and (\getfullref{nland.nonseq}):

\pex<nland>
  \a<seq> John ran to the edge of the cliff and jumped.

  \a<nonseq> John jumped and ran to the edge of the cliff.
\xe

Does this show that the meaning of natrual language connectors differs from their counterpart in formal logic so that linguists must discover a special natural language logic (which is presumably more subtle and complicated than formal logic)? \textcite{griceWilliamJamesLectures1967} \parencite[later published in][]{griceStudiesWayWords1989} argued that this is not the case. Rather, he attributed the difference between (\getfullref{nland.seq}) and (\getfullref{nland.nonseq}) to the fact that natural language productions are used by \emph{speakers} as \emph{utterances} in \emph{conversations} with \emph{hearers}, and such social exchanges are governed by a set of \emph{maxims} that are grounded in rationality. These implicitly shared rationality maxims make it possible for speakers to convey meanings that are more subtle than what the semantics of natural language expressions allows for, and to convey their intended meaning with utterances that are more economical than fully literal expressions would be. Implicatures, and in fact various types of implicatures, are a theoretical tool to describe the difference between semantics and the speaker's meaning.

This general account makes the groundbreaking, even monumental, claims:

\pex<griceclaims>Inference is central in the communication of spekaer's meaning:
\a Speakers choose utterances on the basis of which conversational maxims rational listeners would expect them to follow, while intending to achieve their goals;
\a Listeners comprehend utterances on the basis of what conversational maxims rational communicators would be expected to implement;
\a Inference is constrained by regularities in the conversational domain that give rise to shared assumptions.
\xe

In other words, communication involves a fundamental layer of inference to the best explanation (i.e. non-demonstrative inference). Such inference involves uncertainty, as it can not be guaranteed that the listener will find the best explanation for the speaker's behaviour. However, inference in communication takes place against the background of regularities characteristic of rational action which are implicitly shared by speakers and listeners and make their actions transparent to each other.

Grice sought to explain inference in verbal communication by identifying a set of maxims that rational communicators must be expected to work under. Rational Speech-Act theory (RSA)\footnote{\textcite{frankPredictingPragmaticReasoning2012}, \textcite{goodmanPragmaticLanguageInterpretation2016}, \textcite{goodmanKnowledgeImplicatureModeling2013}. For an introduction, see \textcite{degenRationalSpeechAct2023}, \textcite{erkProbabilisticTurnSemantics2022} and \textcite{scontrasProbabilisticLanguageUnderstanding}.} formalizes this inferential account of communication as a series of interacting probabilistic reasoning processes. Informally, we can outline this intertwined reasoning process as it applies to our example as in (\getfullref{gricersa})\footnote{See also Grice's ``working out schema'' for implicatures in Appendix~\ref{sec:h.-paul-grice}.}:

\pex<gricersa>Mary's and Peter's reasoning based on the assumption that speakers and listeners are rational communicators and follow the maxims as much as possible:
\begin{enumerate}
\item[] \textsc{Mary: } [``pragmatic speaker'' reasoning about a ``pragmatic listener'']
    \begin{enumerate}
  \item I want to make it manifest to Peter that I don't want to drink coffee at 6 pm.
  \item I also want to make it manifest to Peter that I do want something to drink
    and that I do want to be sociable at the party.
  \item I need to choose the utterance that maximizes the probability
    that Peter, assuming that I follow the maxims, can interpret in a way
    that satsifies the maxims and conveys propositions that express my intentions/goals. 
  \end{enumerate}
 \item[] \textsc{Peter: } [``pragmatic listener'' reasoning about a ``pragmatic speaker''
  and how such a speaker differs from a ``literal speaker''
  who in turn reasons about a ``literal listener'']
  \begin{enumerate}
  \item I expect Mary to confirm or reject that Mary wants to drink a cup of coffee at 6 pm.
  \item I expect Mary to be sociable even when declining the offer.
  \item Mary is a rational communicator and follows the maxims. 
  \item Mary's utterance, considering only what is said in it, does not satisfy
    my expectation that she confirms or rejects the idea that Mary wants to drink coffee.
  \item I can see that when considering an implicit premise such as (~\Ref{tab:impicature-compr-coffee} d),
    the conclusion that Mary does not want to drink coffee at this time follows. 
  \item It is manifest to Mary that her utterance yields this conclusion if one accesses
    the (~\Ref{tab:impicature-compr-coffee} d) as implicit premise.
  \item It is manifest that if Mary intends me to access this premise,
    she is justified in believing her utterance to follow the maxims. 
  \item Hence Mary conveyed (~\Ref{tab:impicature-compr-coffee} d) and (~\Ref{tab:impicature-compr-coffee} f) as implicatures. 
  \end{enumerate}
\end{enumerate}
\xe

Notice that this formulation differs considerably from the explanation of the very basic RSA model in \textcite[Chapter 1][]{scontrasProbabilisticLanguageUnderstanding}, which has the following layering: a ``pragmatic listener'' reasons about a ``pragmatic speaker'' who in turn reasons about a ``literal listener''. However, it is unclear how a ``literal listener'' enters the Gricean schema of implicature identification in any useful way.\footnote{%
Therefore it seems to us that either (1) Scontra's et al's ``vanilla RSA model'' deviates from Gricean pragmatics at this point, or (2) their ``vanilla RSA model'' is applicable only to their very simple reference game and needs to be extended when considering more real life communicative behaviour.
} Moreover, what \textcite{scontrasProbabilisticLanguageUnderstanding} and \textcite{tesslerLogicProbabilityPragmatics2022} call a ``pragmatic speaker'' appears to be a misnomer and actually describes a ``naive speaker'', i.e. a speaker who chooses an utterance considering how a hearer who attends only to the literal meaning of the utterance would understand (a ``literal listener'').\footnote{%
  Such a ``naive speaker'' does not differ much from a ``literal speaker'', but \textcite{tesslerLogicProbabilityPragmatics2022} already use the term ``literal speaker'' to a speaker who is producing conclusions to syllogisms literally.
} Such a conception is only ever appropriate for conversations in a language where literal meaning is all there is, and the only questions arising are how appropriate it is to apply meaning M to a given situation S. This is the case in the simple reference game: this language doesn't have syntax, and it has only a handful of words which have extensional meanings. Even the most simple human conversations using natural language are radically different than this, and Grice was one of the first to actually point this out.

\subsubsection{Communication and domain-specific regularities}
\label{sec:comm-doma-spec}

Relevance theory fully adopts the general view of communication outlined in (\getfullref{griceclaims}). However, it deviates from Grice's account in two important respects: on the descriptive side, it claims that the explicit side of communication involves more inference than Grice imagined: in addition to pronoun reference and ambiguity resolution, the meaning of words need to be fine-tuned in context and linguistically unspeficied variables need to be supplied before truth conditions can be evaluated. \parencite[For discussion, see][]{carstonImplicatureExplicatureTruththeoretic1988,carstonThoughtsUtterances2002}

On the explanatory side, \textcite{sperberRelevance1995} seek to find the inference-guiding regularities of the conversational domain not in an implementation agnostic rational analysis (such as in Grice's `working out scheme' for implicatures, \textcite[30--31]{griceStudiesWayWords1989}) but in a psychologically adequate account of mental processes. As we have seen above, this account recognises relevance not only as a property of utterances, but of inputs to cognitive processes in general. It furthermore shows how a modest generalisation about the role of relevance in cognition (the cognitive principles of relevance in \getfullref{cogpor}) has profound effects when the input to cognitive processes happens to be an overtly intentional (i.e. \emph{ostensive}) act: in this case, inference follows a pattern (described in \getfullref{relproc}) that greatly reduces the processing steps needed for the audience to justifiably reach a conclusion about the speaker's intended meaning.

Notice that the inference pattern in the comprehension heuristic in (\getfullref{relproc}) specifies a rule to search through interpretive options (access candidates in order of accessibility, i.e. follow a path of least effort), defines a stopping criterion (check if this interpretation yields cognitive effects as expected) and provides a one-reason framework for decision making (attribute the first interpretation satisfying the stopping criterion to the speaker's intention). In other words, the procedure appears to be an instance of what \textcite[74]{gigerenzerFastFrugalHeuristics2004a} calls a \emph{Take the Best} heuristic: a procedure that is simple, exploits regularities of a specific domain (that ostensive stimuli convey expectations of their own optimal relevance), and allows successfull inference with partial and incomplete knowledge. \smartcite[624--625]{wilsonRelevanceTheory2004} argue that the relevance-theory processing heuristic should be interpreted as such a ``fast and frugal'' cognitive heuristic in Gigerenzer's sense and constitutes a mental module dedicated to the comprehension of ostensive stimuli.

Gigerenzer and colleagues appear to work from the fundamental assumption that probabilistic optimization models are inherently complex and therefore rather unlikely to be able to model mental processes directly. They provide the theoretical gold standard against which the performance of heuristic procedures can be measured, but do not provide insight into how cognitive processes work. If this fundamental assumption is justified, and if we accept Sperber and Wilson's argument that the comprehension procedure they propose is a cognitive heuristic hard-wired into a comprehension module, then it follows that we should apply Bayesian modelling to relevance theory only at the higher level of modelling the conversation exchange. We could then model Mary and Peter's conversational exchange as in (\getfullref{rhrsa}):

\pex<rhrsa>Mary's and Peter's reasoning based on the assumption
that listeners are hard-wired to apply the relevance-theory heuristic:
\begin{enumerate}
\item[] \textsc{Mary: } [``pragmatic speaker'' reasoning about a ``pragmatic listener'']
  \begin{enumerate}
  \item I want to make it manifest to Peter that I don't want to drink coffee at 6 pm.
  \item I also want to make it manifest to Peter that I do want something to drink and that I do want to be sociable at the party.
  \item I need to choose the utterance that maximizes the probability that Peter's first interpretive hypothesis that he tests following a path of least effort will make it more manifest to him that I have these two goals.
  \end{enumerate}
\item[] \textsc{Peter: } [``pragmatic listener'' reasoning about a ``pragmatic speaker'']
  \begin{enumerate}
  \item I expect Mary to confirm or reject that Mary wants to drink a cup of coffee at 6 pm.
  \item I expect Mary to be sociable even when declining the offer.
  \item The most easily accessible interpretation I get is the one in table \Ref{tab:impicature-compr-coffee} and it satisfies my expectations. 
  \end{enumerate}
\end{enumerate}
\xe

Notice how the assumption that the relevance-theoretic comprehension heuristic is hard-wired in the mental architecture of speakers and listeners radically simplifies the layout of the probabilistic modelling of conversation as opposed to (\getfullref{gricersa}).

The comprehension heuristic works on the assumption that different interpretive hypotheses vary in the cost incurred in processing them, and therefore in their order of accessibility. In probabilistic modelling, we can capture this effect by sampling from a set of interpretive hypotheses that are ordered in terms of accessibility. This order is arbitrarily determined by the experimenter to test the effect of various assumptions about the relative cost of constructing interpretive hypotheses. By operating with manually consructed hypothesis sets and accessibility orders we capture the idea that the hypothesis generation stage of non-demonstrative inference is not subject to scientific theorizing, in contrast to the hypothesis evaluation stage.

The comprehension procedure has the effect of making sure that when Peter (the ``pragmatic listener'') samples candidate interpretations, the interpretation satisfying his expectations is among the first few samples drawn, if not the first. In other words, Peter's sampling of candidate interpretations is not based an a uniform draw, but is heavily skewed by the path of least effort. In order to complete our model of the hypothesis evaluation stage of the comprehension procedure/heuristic, we need to look closer at various properties of the comprehension heuristic.

\subsection{Parallel adjustment of context, implicit import and explicit content}
\label{sec:parall-adjustm-cont}

Relevance Theory emphasizes that comprehension is an on-line process:

\begin{quote}
  These subtasks should not be seen as sequentially ordered: the hearer does not FIRST decode the logical form, THEN construct an explicature and select an appropriate context, and THEN derive a range of implicated conclusions. Comprehension is an on-line process, and hypotheses about explicatures, implicated premises and implicated conclusions are developed in parallel against a background of expectations which may be revised or elaborated as the utterance unfolds. \parencite[615]{wilsonRelevanceTheory2004}
\end{quote}

Let us suppose that Peter's language module has just parsed the partial sentence \emph{I don't want to drink\ldots{}} and paired it with the propositional fragment \textsc{Mary does not want to drink\_}. Given the relevance expectation (c) in \Ref{tab:impicature-compr-coffee}, this propositional fragment is enough for Peter that the manifestness of assumption (f) in \Ref{tab:impicature-compr-coffee} is raised enough to form the hypothesis that Mary could have intended to communicate this assumption. Next, Peter's language module has parsed the next chunk of the sentence \emph{I don't want to drink energizing drinks\ldots{}} and paired this chunk with the propositional fragment \textsc{Mary does not want to drink energizing drinks}. This propositional fragment contains the concept \textsc{energizing drink} which we have seen gives Peter easy access to encyclopaedic information along the lines in (d) and (g) in \Ref{tab:impicature-compr-coffee}. This encyclopaedic information, if accessed as contextual assumptions, makes transparent how the proposition (e) in \Ref{tab:impicature-compr-coffee} can provide logical (inferential) support to (f) in \Ref{tab:impicature-compr-coffee}.\footnote{%
  \textcite{bergenPragmaticReasoningSemantic2016} recognise the need for the parallel adjustment of explicit content and implicit import in their model of M-implicature comprehension. As the authors rightly remark, this makes their account close in spirit to relevance theory in this respect.
}

Having unpacked the general idea of the parallel adjustment between implicit import, explicit content and contextual assumptions in on-line processing of utterances, it becomes clear that Wilson and Sperber's rejection of the picture of utterance interpretation as a sequential process (from decoding to explicature to context selection and implicated conclusions) does not advocate for a model of comprehension as a process consisting of completely independend, encapsulated parallel processes. Rather, the idea is that the subtasks of comprehension are interconnected in the sense that early results of linguistic decoding may in the presence of strong expectations about cognitive effects give rise to early hypotheses about intended meaning which may later be substantiated by backwards inference.

This parallel adjustmen of implicit import, explicit conten and conextual assumptions can be modelled as interacting conditional inferences:

\begin{enumerate}
\item ``Bottom-up'' inference: from logical form via contextual assumptions
  and explicit content to implicatures (implicated conclusions).
  \begin{itemize}
  \item For every logical form:
    \begin{itemize}
    \item sample accessible developments of explicit content, 
    \item sample accessible contextual assumptions
    \item yield a probability distribution over implicatures conditional on hypotheses about explicit meaning and contextual assumptions
    \end{itemize}
  \end{itemize}
\item ``Top-down'' inference: from accessible hypotheses about implicit conclusions
  via contextual assumptions to explicatures
  \begin{itemize}
  \item For every accessible implicit conclusion,
    and every logical form:
    \begin{itemize}
    \item sample accessible explicatures
    \item sample accessible contextual assumptions
    \item yield a probability distribution over explicatures conditional on hypotheses about implicatures and contextual assumptions
    \end{itemize}
  \end{itemize}
\end{enumerate}

In this model, the theorist/experimenter supplies a small set of hypothesis triplets <expl, cont, impl> ordered according to the assumed relative order of accessibility. Sampling is done over this ordered set. By changing the content of the hypotheses, as well as by changing the order of relative accessibility, the experimenter can check the results of this inference model against theoretical expectations.

Considering our main example, we propose to operate with the following simplifications:

\begin{itemize}
  \item While every utterance could in principle have several logical forms if the utterance involves structural or lexical ambiguity, this issue does not arise in the example. Therefore we abstract away from this possibility.

  \item While the development of logical forms may involve free enrichment processes as well as filling in of variable values and ad-hoc concept construction for almost every word in the utterance, the example was chosen deliberately with the purpose to limit the amount of logical form enrichment. Therefore we can justifiably limit our attention to a narrow subset of factors. We consider only the possible reference resolution of ``energizing drinks'', and possible on temporal interpretation.
\end{itemize}

The practical implications of the claim that implicit import and explicit content is adjustet in parallel can be seen more clearly in example (\getfullref{supper.li}) from \textcite{wilsonTruthfulnessRelevance2002} \parencite[reprinted in ][pp. 47--83.]{wilsonMeaningRelevance2012}.

\pex<supper>
  \a<aj> \textsc{Alan Jones:} Do you want to join us for supper?

  \a<li> \textsc{Lisa:} No, thanks. I've eaten. \parencite[77]{wilsonMeaningRelevance2012}
\xe

Lisa's utterance \emph{I've eaten} can be understood as a legitimate reason to reject an invitation for supper so long as the temporal reference can be inferentially determined to refer to sometime this evening, and the concept denoted by the verb \emph{eaten} be narrowed to convey the concept \textsc{eat supper} giving the idea that Lisa has eaten something of a kind an quantity that it counts as a full meal. Only when the logical form of Lisa's utterance is developped in this way does the resulting explicature warrant the inference (\getfullref{supperinf}).

\pex<supperinf>
  \a \emph{Implicit premise: } If someone has already eaten a satisfying meal recently then that person has a good reason to reject the offer of a meal
  
  \a \emph{Explicit premise: } Lisa has already eaten supper this evening
  
  \a \emph{Implicit conclusion: } Lisa has a good reason to reject an offer of a meal
\xe

In fact, \citeauthor{wilsonTruthfulnessRelevance2002} go on to show that explicatures typically require a good deal of inferential adjustment to support implicatures to reach an interpretation that satisfies relevance expectations, a point further developed in \textcite{carstonThoughtsUtterances2002}. However, we have purposefully chosen an example where the amount of inferential adjustment to the explicature is arguably minimal, in order to focus more clearly on the implicatures.

What these considerations show is that there is no need for a stage in processing where a representation that is more developed than the logical form of the utterance, but less developed than the explicature (the intended proposition) plays a role. In other words, there is no level like the Gricean concept of \emph{What is said} or a \emph{Literal meaning} of the utterance that plays a role in utterance interpretation even though it may need to be adjusted, or even rejected and replaced, in later stages of interpretation. So when we develop a probabilistic model of inference in utterance comprehension, we must make sure that neither the speaker nor the listener is reasoning about a ``literal listener'' or a ``literal speaker'', as it is standard practice in RSA work.

\subsection{The central role of relevance expectations and their fine-tuning}
\label{sec:centr-role-relev}

\subsubsection{Relevance expectations, strong and weak implicatures}
\label{sec:relev-expect-strong}

Recall that the stopping criterion of the comprehension heuristic makes reference to the audience's expectations of relevance. Expectations of relevance are raised in general terms by the communicative principle of relevance. However, for this general presumption to be operative in the comprehension heuristic, it must be fine-tuned with reference to the audience's actual expectations at the particular moment. This is how (c) differs from (b) in table \Ref{tab:impicature-compr-coffee}.

The example we are discussing involves an indirect answer to a specific question. In this case the audience has a very specific expectation of cognitive effects, indeed: that the communicator will convey some proposition that either affirms the propositional content of the yes-no question (by contextual strenghtening), or negates it (by contextual contradiction and elimination). It is this very specific expectation that drives the parallel adjustment of explicit content, implicit import and context by backwards inference.

However, not every utterance raises specific expectations of relevance. Consider (\getfullref{noexp}): Mary is walking in town and a passer-by asks her the time.

\pex<noexp>
  \a<pb> \emph{Passer-by: } Excuse me, what's the time?

  \a<ma> \emph{Mary: } It's 2:30 pm.

  \a<me> [Proposition conveyed by Mary: ] \textsc{The time at present is 2:30 pm}
\xe

What is the passer-by's expectation of relevance? Mary has no way to know, other than that a true answer (to the best of Mary's capabilities) will be optimally relevant to him. Mary can be sure that when she communicates a true proposition along the lines of (\getfullref{noexp.me}), her utterance will achieve some cognitive effects in the hearer's mind, but she can not foresee of what nature they may be. In such a situation, it is actually not possible for Mary to convey any implicatures, and her utterance is relevant to the audience in virtue of its explicature alone.

Since relevance consists of the balance of cognitive effects and processing effort, relevance expectations may relate to these factors individually or jointly: an audience may at times have rather specific expectations about the type or content of cognitive effects that would be relevant to him. At other times, he may have more general expectations about the amount of cognitive effects to be achieved. On yet other occassions, the audience's relevance expectations may relate to the effort side. \textcite[237]{sperberRelevance1995} discuss the example of Gustave Flaubert commenting on the poet Leconte de Lisle:

\pex<flb> 
  His ink is pale. (Son encre est pale.) [Sperber and Wilson's example 108]
\xe

The audience's expectations of relevance before processing this utterance are specific enough in the sense that it expects the utterance to be relevant by way of giving some indication of Flaubert's opinion on de Lisle's poetry. But there is no strong indication of what direction these indications may take. In other words, the audience is processing the utterance with the expectation that its processing effort will be offset by enough cognitive effects to be worth the effort (if not more), but it has no prior expectations of which type of content they may have. \citeauthor{sperberRelevance1995}'s analysis is worth quoting in full:

\begin{quote}
  A strictly literal construal of this utterance is clearly ruled out: it is hard to see what relevance could attach to knowing the colour of a poet's ink. Nor is there any obvious strong implicature. The only way of establishing the relevance of this utterance is to look for a wide range of very weak implicatures. This requires several extensions of the context. In the most accessible context of encyclopaedic information about ink and handwriting, most implications are irrelevant: after all, Leconte de Lisle's poetry is read not in his handwriting but in print; the only clear implicature in this first context is that he has the character of a man who would use pale ink. Some other implications -- that Leconte de Lisle's writing lacks contrasts, that it may fade -- have further relevant implications in a context to which has been added the premise that what is true of his handwriting is true of his style. Someone who knows little of Leconte de Lisle's work might conclude, for example, that there is something weak about his poetry, that his writings will not last, that he does not put his whole heart into his work, and so on. Someone who has a deeper acquaintance with the poet would be able to construe the criticism in much more detailed and pointed ways. The resulting interpretation, with its characteristic poetic effect, owes a lot simultanewouly to Flaubert, for foreseeing how it might go, and to the reader, for actually constructing it. \parencite[237]{sperberRelevance1995}
\end{quote}

The upshot of this is that the composition of the content of the communicator's informative intention -- the speaker's meaning -- varies in relation to the nature of the extant relevance expectations in the communication situation. Strong implicatures can only be communicated in situations where the relevance expectations are strong and specific enough to guide the audience in backwards inference. As we have seen, some weak implicatures are communicated in this case as well to give a rational for speaking implicitly at all. When the relevance expectations are non-specific and weak, the communicator can not convey implicatures, and the utterance will be relevant in terms of its explicature alone. When relevance expectations are strong but non-specific, the speaker's meaning may consist of a fuzzy set of lots of weak implicatures (and presumably explicatures giving access to these implicatures as well). This is summarised in table \Ref{tab:relexpmeaning}. Needless to say that these categories are not exclusive ones; rather, many utterances in verbal communication will fall somewhere in between the space plotted out by these examples, with respect to the strenght and specifity of the relevance expectations raised and the composition of the content of the speaker's informative intention.\footnote{%
  Recall furthermore that communication does not aim at achieving identity of the audience's thoughts and speaker meaning; rather, communicators aim at a relevant overlap of thoughts.
}

\begin{table}
  \centering
  \begin{tabularx}{\linewidth}{lXl}
  \toprule
    Relevance expectations & Speaker's meaning & Example \\ 
  \midrule
    strong, specific & strong implicatures, explicature(s) and some weak implicatures & (\getfullref{impl.implmary}) \\
    strong, non-specific & weak implicatures and explicature(s) & (\getfullref{flb}) \\
    weak, non-specific & explicature & (\getfullref{noexp})
\end{tabularx}
  \caption[relevance expectations and speaker's meaning]{The nature of relevance expectations and the composition of the content of the speaker's informative intention}
  \label{tab:relexpmeaning}
\end{table}

\subsubsection{Relevance and the utility of utterances}
\label{sec:relev-util-utter}

Rational speech act theory models speakers as Bayesian decision makers who choose their utterance optimizing its expected \emph{utility} to get the audience to comprehend the speaker's meaning. Much previous work in RSA has focussed on communication situations where the utterance is relevant solely in virtue of its explicature: reference games \parencite{frankPredictingPragmaticReasoning2012} or the semantics and pragmatics of scalar expressions\footnote{%
  These are usually analysed as carrying \emph{scalar implicatures} \parencite{hornNewTaxonomyPragmatica}, but see \textcite{carstonQuantityMaximsGeneralised1995} for arguments that the pragmatic contributions of scalar terms affect explicatures rather than implicatures.
} \parencite{goodmanKnowledgeImplicatureModeling2013}\footnote{%
See \textcite{bergenPragmaticReasoningSemantic2016} for a noticeable exception to the trend of focussing narrowly on scalar inferences. \citeauthor{bergenPragmaticReasoningSemantic2016} study so-called M-implicatures, where the speaker uses a more elaborate, more expensive to process way of expression than necessary (imagine Mary answering Peter: \emph{I do not ingest melatonin receptor blockers less than 6 hours before my regular bed time}, which would not only allow Peter to infer that Mary does not want coffee now, but more likely than not convey more implications such as that Mary may be a nerd that is difficult to deal with). Still, this type of implicatures is only a subset of conversational implicatures, and so the question arises how well their model might generalise to the full range of implicatures.
}. In these instances, the utility of the utterance was seen as a reflection of its power to guide the audience to an informative interpretation, and this in turn was measured in the expression's information-theoretic \emph{negative surprisal}. However, in most communication events the speaker's meaning is not comprised of one proposition alone, but several: an explicature and several implicatures, each of which the communicator intends to make more manifest to the audience to varying degrees. How should the utility function be defined in this case?

Consider again what it means to say that Mary communicates the explicature in table \Ref{tab:impicature-compr-coffee} (e) and the implicatures in table \Ref{tab:impicature-compr-coffee} (d), (f), (g), (h) to Peter. According to the definition of the communicator's informative intention in (\getfullref{intention.informative}), it means that Mary's utterance alters the manifestness of each of these propositions for Peter. This is paramount to saying that Mary changes Peter's probability distribution over these propositions from a prior value to a posterior value. The utility of Mary's utterance can then be measured in terms of the divergence between Peter's prior probability distribution over the set of these propositions to Peter's posterior probability distribution over this set. In information-theoretic terms, this is the Kullback-Leibler divergence between the two probability distributions in question.\footnote{%
\textcite[p. 583]{tesslerLogicProbabilityPragmatics2022} deal with a similar problem when they consider a speaker communicating her conclusion of a syllogistic argument to an audience. The conclusion of a syllogistic argument is comprised of a set of states, and the utility of a conclusion can be ``quantified by how well it would align the listener's beliefs with those of the reaoner (speaker).''
}

There is an other way to conceptualise the utility of an utterance. The utility of an utterance can be understood as the efficiency with which it aligns the audience's interpretation with the communicator's informative intention, in other words: its relevance (in the relevance-theoretic sense). Would it then not be simpler to directly calculate with estimates of cognitive effects vs. processing effort measures? For example, could we not make rough estimates about the number of cognitive effects and approximate the processing effort to the number of inference steps and/or context extension steps? Such estimates would require us to make significant assumptions about actual cognitive implementation which are most likely wrong. In addressing these matters in some detail, \textcite[123--132]{sperberRelevance1995} conclude that relevance is a non-representational property:

\begin{quote}
  Mental effects and effort are non-representational properties of mental processes. Relevance, which is a function of effect and effort, is a non-representational property too. That is, relevance is a property which need not be represented, let alone computed, in order to be achieved. When it is represented, it is represented in terms of comparative judgements and gross absolute judgements, (e.g. `irrelevant', `weakly relevant', `very relevant'), but not in terms of fine absolute judgements, i.e. quantitative ones. \parencite[132]{sperberRelevance1995}
\end{quote}

We submit that the Kullback-Leibler divergence between the audience's prior and posterior probability distributions over the set of communicated assumptions is a viable way to conceptualise comparative judgements of relevance in an indirect way, which is suitable for developing probabilistic models of utterance interpretation while avoiding the trap of trying to quantify non-representational properties.

\subsubsection{Relevance expectations as a stopping criterion for the heuristic procedure}
\label{sec:relev-expect-as}

A cognitive heuristic procedure is triggered by the recognition of an input stimulus affecting the domain of the procedure. In the case of communicative stimuli, this will mean stimuli that make a layered set of communicative and informative intentions overt. The procedure must also stop at some point, so a heuristic procedure must also specify a stopping criterion. In the case of communication, the stopping criterion is the satisfaction of the audience's expectations of relevance \emph{effective in the present communication situation}.

We have already seen in section \Ref{sec:relev-expect-strong} that relevance expectations may be of different types according to the different factors of cognitive effects and processing effort that make up relevance:

\pex<relexptypes>
  \a<tp1> Expectations about specific content of cognitive effects

  \a<tp2> Expectations about the kind of information that cognitive effects may convey

  \a<tp3> Expectations about the relative size of sets of cognitive effects

  \a<tp4> Expectations about the level of processing effort expected of the audience
\xe

Moreover,
we have seen in table \Ref{tab:impicature-compr-coffee} that specific expectations of relevance are adopted on the basis of, and with the purpose of making concrete, the claim of optimal relevance which according to the communicative principle of relevance is communicated in every instance of ostensive communication.\footnote{More precisely,
  in every instance of communication involving stimuli making
  a communicative and informative intention overt.
  There may be instances of ostensive behaviour that do not necessarily
  make a communicative intention overt, see [REF: find Sperber \& Wilson 2024].
} \textcite[107--142]{ungerGenreRelevanceGlobal2006} points out that the inference processes involved in this step from the general claim of optimal relevance to the fine-tuning of relevance expectations may have profound impact on the processing of complex ostensive stimuli such as longer texts. However, the details of this sub-process of comprehension is rarely discussed in detail in the literature. The rational for this is presumably due to the assymetry of responsibilities of communicators and audience: whereas communicators must make sure that they choose utterances that are optimally relevant to their audience, audiences are licensed to presume that this is the case. This means that it is speakers who must evaluate what may be relevant to their hearers, whereas hearers do not need to double check what conclusions the speaker has come up with in this process.

Recall at this point that Rational Speech Act theory addresses both the speaker's and the listener's perspectives in a symmetrical way: the speaker reasons about how a `literal listener' will understand an utterance under consideration, and how a `pragmatic listener' will deviate from this. Likewise, the listener reasons about what a `literal speaker' might have said to communicate a certain piece of information, and how a `pragmatic speaker' deviates from this choice (see section~\Ref{sec:comm-infer-under}). A corresponding probabilistic model of relevance theory capturing both communicator's and audiences' perspectives would need to capture the assymetry of these communicative roles more clearly, making the description of the communicator's inferences about the audience's state of mind with respect to relevance much more complex. Clearly, the computational modelling of how communicators evaluate (and influence) what is relevant to other individuals is an important challenge to address. But in this paper we follow the path of classical relevance theory by concentrating the effort of explanation at the comprehension phase of communication.

The question arising at this point is how to handle relevance expectations in a probabilistic computational model. Expectations of relevance relating to cognitive effects can be expressed or paraphrased as questions \parencite[143--173]{ungerGenreRelevanceGlobal2006} which are often called \emph{questions under discussion} \parencite{BeaverRoberts2017}. These questions are about what may be infered or concluded on the basis of information accessible in the context and provided in the utterance. In logic programming, inference is treated as answering the question whether a certain conclusion can be traced back to the ground facts provided by the knowledge base section of the program.\footnote{%
  See \textcite{flach1994simply} for an accessible introduction to logic programming in Prolog. A fully interactive online edition can be found at the URL \url{https://book.simply-logical.space/src/simply-logical.html}.
}
In probabilistic logic programming, the answer to this question is not a binary choice between \textsc{true} or \textsc{false}, but a probability value indicating how likely it is that the queried proposition is true. In the probabilistic programming language ProbLog (version 2)\footnote{%
  See \textcite{fierensInferenceLearningProbabilistic2015} for thorough and accessible introduction to probabilistic programming in ProbLog.
}, which we will be using in our implementation, inference processes answering the question as to how probable (i.e., how manifest) certain expected cognitive effects are, are triggered by using the \verb+query()+ predicate at the end of the program (see the code listings in the appendices).

Notice that this approach to checking relevance expectations by querying the program for answers to their corresponding questions first and foremost deals with specific relevance expectations as to content of cognitive effects (type \getfullref{relexptypes.tp1}). By passing a whole set of queries about various questions under discussion we can also cover other types of relevance expectations of the types (\getfullref{relexptypes.tp2}) and (\getfullref{relexptypes.tp3}). These are the types of relevance expectations arising in our main example.\footnote{%
  We cannot directly model relevance expectations relating to processing effort alone, i.e. expectations of type (\getfullref{relexptypes.tp4}). Expectations of this kind are arguable raised in utterances communicating only a wide range of weak implicatures, as in example (\getfullref{flb}). Notice, however, that there is a close correspondence between relevance expectations of type (\getfullref{relexptypes.tp3}) and (\getfullref{relexptypes.tp4}): when the communicator promises the audience that they can satisfy their relevance expectations by investing a relatively high amount of processing effort, it follows that checking whether this promise is true amounts to processing a sizeable set of cognitive effects. This means that we can indirectly model relevance expectations of type (\getfullref{relexptypes.tp4}) by transforming them hypothetically as instances of (\getfullref{relexptypes.tp3}).
}

\section{Developing a probabilistic computational model of implicature comprehension}
\label{sec:devel-prob-model}

In this section we describe our implementation of a probabilistic computational model of implicature comprehension according to relevance theory in the probabilistic programming language ProbLog \parencite{fierensInferenceLearningProbabilistic2015}.

\subsection{Drawing a knowledge graph for encyclopaedic knowledge and context selection}
\label{sec:knowl-graph-encycl}

\begin{figure}
  \centering

  \begin{tikzpicture}
    \node (cof) at (0,4cm) {coffee};
    \node (tea) at (0,3cm) {tea};
    \node (wat) at (0,2cm) {water};
    \node (pep) at (0,1cm) {peppermintTea};
    \node (fru) at (0,0) {fruitTea};
    \node (cok) at (0,-1cm) {coke};
    \node (red) at (0,-2cm) {redBull};
    \node (pro) at (0,-3cm) {proteinShake};
    
    \node (tbd) at (4.3cm,3cm) {
      \begin{minipage}{3cm}
        \textbf{t}iredness-\textbf{b}locking \textbf{d}rink
      \end{minipage}};
    \node (ped) at (4.3cm,1cm) {
      \begin{minipage}{3cm}
        \textbf{p}erformance-\textbf{e}nhancing \textbf{d}rink
      \end{minipage}};
    \node (mgd) at (4.3cm,-1cm) {
      \begin{minipage}{3cm}
        \textbf{m}uscle-\textbf{g}rowing \textbf{d}rink
      \end{minipage}};
    
    \node (par) at (7.3cm,-3cm) {party drink};

    \node (drk) at (10cm,0) {drink};

    \graph [edge quotes={fill=white, inner sep=1pt}] {
      (cof) ->[line width=0.9pt] (tbd) ;
      (cof) ->[line width=0.2pt] (ped) ;
      (cof) ->[line width=0.9pt] (par) ;

      (tea) ->[line width=0.85pt, green] (tbd) ;
      (tea) ->[line width=0.2, green] (ped) ;
      (tea) ->[line width=0.9, green] (par) ;

      (wat) ->[line width=0.9pt, blue] (par) ;
      (wat) ->[line width=0.1pt, blue] (ped) ;
      
      (tbd) ->[line width=0.9pt] (drk) ;
      (ped) ->[line width=0.7pt] (drk) ;
      (mgd) ->[line width=0.7pt] (drk) ;

      (pep) ->[line width=0.4pt, orange] (tbd) ;
      (pep) ->[line width=0.9pt, orange] (par) ;

      (fru) ->[line width=0.8pt, red] (par) ;
      (fru) ->[line width=0.1pt, red] (tbd) ;
      (fru) ->[line width=0.1pt, red] (ped) ;

      (cok) ->[line width=0.8pt, brown] (tbd) ;
      (cok) ->[line width=0.6pt, brown] (ped) ;
      (cok) ->[line width=0.8pt, brown] (par) ;

      (red) ->[line width=0.9pt, pink] (ped) ;
      (red) ->[line width=0.8pt, pink] (tbd) ;
      (red) ->[line width=0.3pt, pink] (par) ;

      (pro) ->[line width=0.87pt, yellow] (mgd) ;
      (pro) ->[line width=0.87pt, yellow] (ped) ;
      (pro) ->[line width=0.1pt, yellow] (par) ;

      (par) ->[line width=1pt]  (drk) ;

      };
  \end{tikzpicture}

  \caption[Graph hypothesis 1]{Graph representation of encyclopaedic knowledge in hypothesis 1}
  \label{fig:knowledgegraph_drinks}
\end{figure}
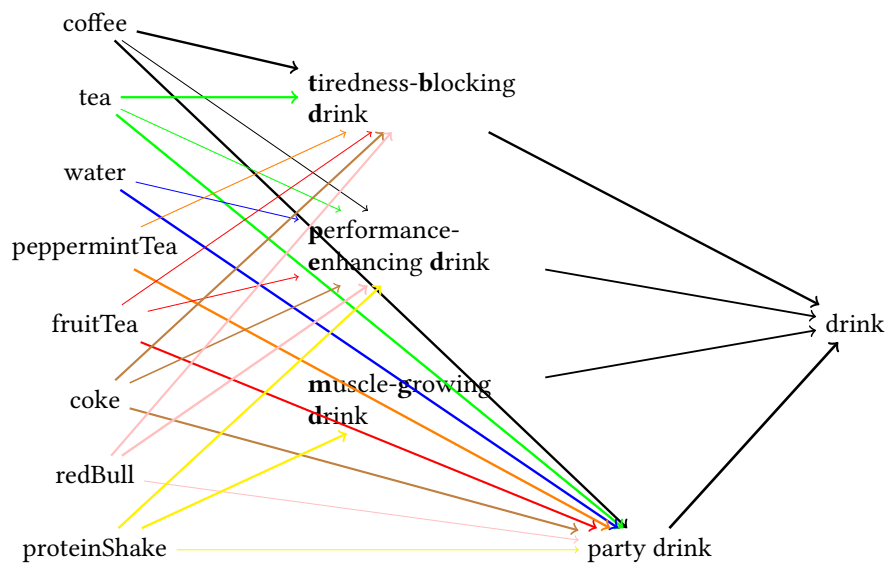

Encyclopaedic knowledge linked to the concepts triggered by the words in the sentence conveyed in the utterance provides the starting point for context selection. In our example (\getfullref{impl}), we are dealing mainly with encyclopaedic knowledge relating to \emph{coffee} and various other beverages that may typically be offered at parties, and their perceived properties (e.g. whether they block tiredness, enhance performance, etc.). This kind of knowledge can be effectively encoded in a graph structure. Inference over such a structure amounts to the query whether a path exists from a starting node A to a goal node B. ProbLog allows to encode probabilistic graphs, that is, the information that there is an edge (a link) between two nodes A and B can be annotated with a probability value (a number between 0 and 1). Figure \Ref{fig:knowledgegraph_drinks} shows a schematic representation of the knowledge graph we use as a starting point. The thickness of the individual arrows linking nodes vary according to the probability values encoded. Lines 4--30 in the code listing in appendix \Ref{sec:code-list-optim} show the actual implementation of this graph.\footnote{%
  For a more detailed discussion of the use of probabilistic graphs for concept and word meaning, see \textcite{erkHowMarryStar2023}. Notice, however, that \citeauthor{erkHowMarryStar2023} tend to not making a clear distinction between word or concept meaning and encyclopaedic knowledge linked to to concepts.
}

\subsection{Linking words and concepts by inference}
\label{sec:infer-link-words}

One task of utterance interpretation is the development of the logical form of the sentence uttered. This includes the question of which concepts are conveyed by words and phrases. In our example, a central question is what concept is conveyed by the compound word \emph{energizing drink}. In other words, we need to establish a link between the word \emph{energizing drink} and certain nodes in the knowledge graph representing encyclopaedic knowledge. We implement this by writing a disjunction of inference rules of the kind given in (\getfullref{endrinkrules}):

\pex<endrinkrules>
X is an energizing drink, \ldots{}
\begin{itemize}
\item if there is a path from X to the concept \textsc{tiredness-blocking drink}, OR
\item if there is a path from X to the concept \textsc{performance-enhancing drink}, OR
\item if there is a path from X to the concept \textsc{muscle-growing drink}. 
\end{itemize}
\xe

Each disjunctive rule is assigned a probability value. See lines 40--45 in the code listing in appendix \Ref{sec:code-list-optim}.

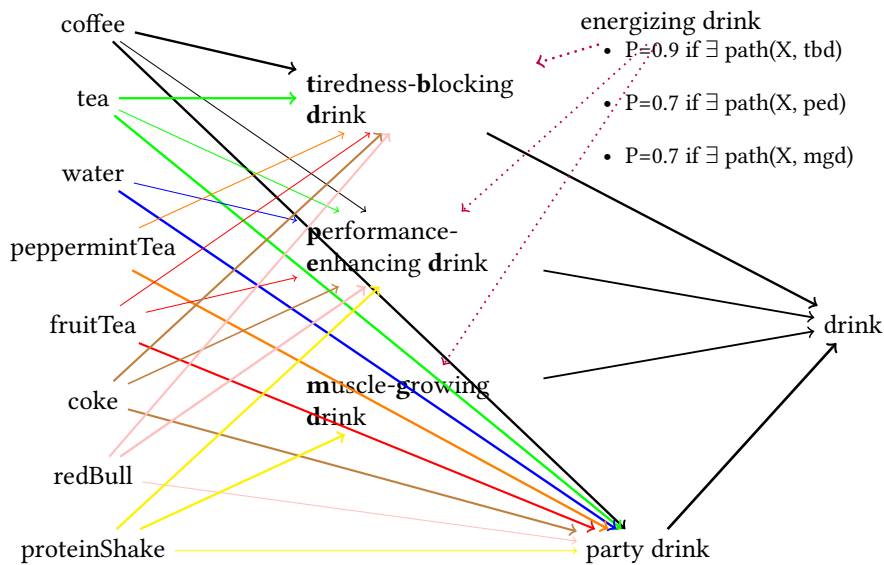
\begin{figure}
  \centering

  \begin{tikzpicture}
    \node (cof) at (0,4cm) {coffee};
    \node (tea) at (0,3cm) {tea};
    \node (wat) at (0,2cm) {water};
    \node (pep) at (0,1cm) {peppermintTea};
    \node (fru) at (0,0) {fruitTea};
    \node (cok) at (0,-1cm) {coke};
    \node (red) at (0,-2cm) {redBull};
    \node (pro) at (0,-3cm) {proteinShake};
    
    \node (tbd) at (4.3cm,3cm) {
      \begin{minipage}{3cm}
        \textbf{t}iredness-\textbf{b}locking \textbf{d}rink
      \end{minipage}};
    \node (ped) at (4.3cm,1cm) {
      \begin{minipage}{3cm}
        \textbf{p}erformance-\textbf{e}nhancing \textbf{d}rink
      \end{minipage}};
    \node (mgd) at (4.3cm,-1cm) {
      \begin{minipage}{3cm}
        \textbf{m}uscle-\textbf{g}rowing \textbf{d}rink
      \end{minipage}};
    
    \node (edr) at (7.6cm,4cm) {energizing drink};
    \node (par) at (7.3cm,-3cm) {party drink};

    \node (edf) at (6cm,3.9cm) [anchor=north west] {
      \begin{minipage}{5cm}
        \begin{itemize}
        \item {\small P=0.9 if $\exists$ path(X, tbd)}
        \item {\small P=0.7 if $\exists$ path(X, ped)}
        \item {\small P=0.7 if $\exists$ path(X, mgd)}
        \end{itemize}

      \end{minipage}};
    
    \node (drk) at (10cm,0) {drink};

    \graph [edge quotes={fill=white, inner sep=1pt}] {
      (cof) ->[line width=0.9pt] (tbd) ;
      (cof) ->[line width=0.2pt] (ped) ;
      (cof) ->[line width=0.9pt] (par) ;

      (tea) ->[line width=0.85pt, green] (tbd) ;
      (tea) ->[line width=0.2, green] (ped) ;
      (tea) ->[line width=0.9, green] (par) ;

      (wat) ->[line width=0.9pt, blue] (par) ;
      (wat) ->[line width=0.1pt, blue] (ped) ;
      
      (tbd) ->[line width=0.9pt] (drk) ;
      (ped) ->[line width=0.7pt] (drk) ;
      (mgd) ->[line width=0.7pt] (drk) ;

      (pep) ->[line width=0.4pt, orange] (tbd) ;
      (pep) ->[line width=0.9pt, orange] (par) ;

      (fru) ->[line width=0.8pt, red] (par) ;
      (fru) ->[line width=0.1pt, red] (tbd) ;
      (fru) ->[line width=0.1pt, red] (ped) ;

      (cok) ->[line width=0.8pt, brown] (tbd) ;
      (cok) ->[line width=0.6pt, brown] (ped) ;
      (cok) ->[line width=0.8pt, brown] (par) ;

      (red) ->[line width=0.9pt, pink] (ped) ;
      (red) ->[line width=0.8pt, pink] (tbd) ;
      (red) ->[line width=0.3pt, pink] (par) ;

      (pro) ->[line width=0.87pt, yellow] (mgd) ;
      (pro) ->[line width=0.87pt, yellow] (ped) ;
      (pro) ->[line width=0.1pt, yellow] (par) ;

      (edr) ->[line width=0.9pt, dotted, purple] (tbd);
      (edr) ->[line width=0.7pt, dotted, purple] (ped);
      (edr) ->[line width=0.7pt, dotted, purple] (mgd);
      (par) ->[line width=1pt]  (drk) ;

      };
  \end{tikzpicture}
  
  \caption[Linking words and concepts]{Linking word meaning to concepts in the encyclopaedia}
  \label{fig:link_words_concepts}
\end{figure}

These rules allow us to formulate queries of the kind \emph{is coffee an energizing drink?} or \emph{is tea an energizing drink?} and receive probabilistic answers. However, more interesting than begin able to execute such queries in isolation is the fact that we can use the probabilistic paths that such inferences open up in a net of wider inferences linking the description of Mary's utterance to its explicature, and thereby also to its implicatures.\footnote{%
  In this way we approximate ad-hoc concept identification to disambiguation, i.e. a process of choosing among pre-existing concepts represented in the mind of the hearer. While ad-hoc concept resolution may in some cases come down to choosing among pre-existing concepts in this way, ad-hoc concepts are typically concepts created on the spot for the given occassion and not persistently stored in memory (\textcite{sperberMappingMentalPublic1998}, also published as chapter 2 of \textcite[31--46]{wilsonMeaningRelevance2012}). Since our main purpose is to model implicature comprehension, we do not address this issue. Notice that this limitation is also inherent in \textcite{erkHowMarryStar2023}, who focus on modelling concepts that individuals persistently mentally represent.
} Figure \Ref{fig:link_words_concepts} illustrates these links.

\subsection{Adjusting explicatures and implicatures in a Bayesian network}
\label{sec:bayes-netw-adjust}

The parallel adjustment process of implicatures and explicatures can be modelled in a Bayesian network as indicated in figure \Ref{fig:implexplbayesnw}. In our example, the implicature \textsc{Mary does not want coffee} depends on the one hand on the likelihood of the explicature \textsc{Mary does not drink tiredness-blocking drinks after 6PM} being true, and on the likelihood of the implicit premise \textsc{Coffee is a tiredness-blocking drink} being true, on the other. Moreover, the explicature \textsc{Mary does not drink tiredness-blocking drinks after 6PM} depends on the likelihood of Mary having said ``I do not drink energizing drinks at this time of the day'' and on the likelihood of the word ``energizing drink'' communicating the ad-hoc concept \textsc{tiredness-blocking drink}.\footnote{%
  We are abstracting away from other factors entering into establishing the link between the sentence uttered and its conceptual representation.
}

\begin{figure}
  \centering
  
  \begin{tikzpicture}
    \path
    (0cm,4cm) node[draw] (says) {
      \begin{minipage}{4cm}
        {\small Mary says S = 
          \emph{I don't drink energizing-drinks\ldots{}}}
      \end{minipage}}
    -- (6cm,4cm) node[draw] (tbd) {
      \begin{minipage}{5cm}
        \begin{center}
          {\small \textsc{Energizing-drink*} $\rightarrow$ \textsc{tiredness-blocking drink}}          
        \end{center}
      \end{minipage}}
    -- (1.5cm,2cm) node[draw] (exp) {{\small \textsc{Mary does not drink tiredness-blocking drinks\ldots{}}}}
    -- (8cm,2cm) node[draw] (xis) {X is ED*}
    -- (5cm,0cm) node[draw] (ico) {Mary does not want X};
    \draw (says) -- (exp);
    \draw (tbd) -- (exp);
    \draw (exp) -- (ico);
    \draw (xis) -- (ico);
    \node at (-0.5cm,3.4cm) [anchor=north east] {P(S)};
    \node at (4.5cm,3.4cm) [anchor=north west] {P(A)};
    \node at (1.5cm,1.5cm) [anchor=north east] {P(E|S,A)};
    \node at (7.8cm,1.7cm) [anchor=north west] {P(Ca)};
    \node at (5cm,-1cm) [anchor=south] {P(I|E,Ca)};
    \node (legenda) at (-2cm,-1cm) [anchor=north west] {
      \begin{minipage}{1.0\linewidth}
        {\small
          S: description of the \textbf{S}entence uttered \\
          A: pragmatic concept \textbf{A}djustment \\
          E: \textbf{E}xplicature \\
          Ca: \textbf{C}ontextual \textbf{a}ssumption; implicit premise \\
          I: \textbf{I}mplicature; implicated conclusion      }
      \end{minipage}
    };
  \end{tikzpicture}
  
  \caption[implicature-explicature as Bayesian network]{Parallel adjustment of implicatures and explicatures as a Bayesian network}
  \label{fig:implexplbayesnw}
\end{figure}

\subsection{Evaluating interpretive hypotheses}
\label{sec:eval-interpr-hypoth}

We have seen in section \Ref{sec:proc-overtly-intentional} that the communicative principle of relevance justifies a heuristic procedure that stops at the first easily accessible interpretive hypothesis that satisfies the audience's relevance expectations. The communicative principle of relevance entails that the interpretation found by this procedure is optimally relevant to the audience. In the previous section we have argued that in a probabilistic formalisation of relevance theory, we should capture the effects of relevance indirectly by considering the divergence of the probability distributions over sets of manifest propositions before and after processing utterances. The question arising at this point is what this means for modelling the process of finding the winning interpretive hypothesis.

In order to answer this question, we will procede as follows: we will construct various interpretive hypotheses \emph{Int 1, Int 2, Int 3, Int 4, Int 5} differing in specific properties. We will then discuss what effects these properties have on the KL-divergence between the audience's prior estimates of probability/manifestness of propositions relating to the expected cognitive effects and the change in this probability distribution affected by the utterance under the respective interpretive hypothesis. We hope to show that when two interpretive hypotheses differing only in this KL measure, the hypothesis with the highest KL measure is the one that relevance theory predicts to be the optimally relevant one. However, not all interpretive hypotheses differ in their KL measure. We demonstrate that in such cases a comparison of the size of the Herbrand base of the probabilistic program -- that is, the set of facts listed in the knowledge-base section of the program -- allows to pick out the less costly to process hypothesis.\footnote{%
  Apart from cases of genuine amgiuity in interpretation or accidental relevance. An example for the former may be Mozart's comment to his rival Salieri: \emph{I didn't know that music like this is possible.} () An example for the latter case may be Margret Thatcher's comment \emph{I always treat other people's money as if it were my own.} Relevance theory makes no prediction about how audiences settle on an interpretation in such cases.
}

\pex<inthypprop> Interpretive hypotheses and their properties
  \a<int1> \emph{Int 1:} Optimally relevant interpretation: conveys a strong and various weaker implicatures satisfying relevance expectations about what beverage Mary is likely not to drink in addition to coffee. No superfluous context. Implementation provided in appendix \Ref{sec:code-list-optim}.
  
  \a<int2> \emph{Int 2:} Relevance expectations as in Int 1, but only enough contextual assumptions to satisfy only one implicit conclusion about Mary not drinking coffee. Does not satisfy extant relevance expectations.
  
  \a<int3> \emph{Int 3:} Allows the same inferences to be drawn as in Int 1, satisfying the same relevance expectations as Int 1, but has additional superfluous encyclopaedic knowledge. This is supposed to be less relevant than Int 1 due to increased processing effort.
  
  \a<int4> \emph{Int 4:} Contains the same encyclopaedic knowledge about beverages at parties as Int 1, but adds another knowledge graph that allows inferences about whether Mary may be a particularly health-conscious person when not drinking tiredness-blocking drinks after 6PM. However, relevance expectations remain the same as in Int 1, so the extension of the space of possible available inferences does not improve relevance expectation satisfaction.
  
  \a<int5> \emph{Int 5:} Same as Int 3: but now the relevance expectations include questions about Mary's health-consciousness (and what this may mean not only for drinks, but perhaps also for foods that Peter plans to offer later at the party).
\xe

\subsubsection{Interpretive hypothesis Int 1}
\label{sec:interpr-hypoth-int-1}

This interpretive hypothesis assumes that it is highly relevant to Peter to know whether Mary wants to accept his offer of a cup of coffee, and that it would be relevant to him to know that in the case Mary declines coffee, what alternative he may be able to offer with less chance of refusal. We model these relevance expectations as a set of queries of the form \verb+wantsNotDrink(mary,X)+ which can be seen as a placeholder for propostions of the form \textsc{Mary does not want to drink X}, where X is a variable standing for the various beverages that Peter happens to offer. We assume that this list includes coffee, coke, fruit tea, peppermint tea, tea, water, and also some untypical beverages that Peter may be able to offer: protein shakes and Red Bull (energy drinks). We further assume that Peter has no particular knowledge of Mary's preferences, so prior to Mary's utterance, none of the assumptions that Mary does not want to drink any of the available beverages are highly manifest to him, the likelihood that Peter mentally represents any of them as a true proposition is about chance level, perhaps assuming that people are slightly more likely to drink coffee than peppermint tea of fruit tea at 6PM, and less likely to want to drink a non-typical party drink. This is captured in the probability distribution in the column P (prior).

Mary's utterance provides Peter with the evidence that she does not energizing drinks after 6PM. Passing this evidence to the ProbLog program listed in \Ref{sec:code-list-optim} means setting the probability that Mary does not want to drink any X that is connected to the category \textsc{energizing drink*} to true (P=1.0). Given the encyclopaedic knowledge graphs described in \Ref{sec:knowl-graph-encycl} and \Ref{sec:infer-link-words}, this will change the likelihood that Peter is able to mentally represent the respective propositions in the ways indicated in the column P (after). The divergence between the probability distribution P (prior) and P (after) is calculated in the column KLD. The KL measure in this column is calculated with the \verb+kl_div+ function of the Python library \verb+scipy.special+. The function is defined as follows:
  
\begin{displaymath}
  kl\_div(x,y) =
  \begin{cases}
    x * log(x/y) - x + y & x>y, y>0 \\
    y & x=0, y \geq 0 \\
    \infty & otherwise
  \end{cases}
\end{displaymath}

\medskip\noindent
\begin{tabularx}{\linewidth}{lXXX}
\toprule
  Assumption (thought) & P (prior) & P (after) & KLD \\
\midrule
  wantsNotDrink(mary,coffee):	& 0.40	& 0.9208    	& 0.1872 \\
  wantsNotDrink(mary,coke):	& 0.50	& 0.92      	& 0.1151 \\
  wantsNotDrink(mary,fruitTea):	& 0.60	& 0.0199    	& 1.4636 \\
  wantsNotDrink(mary,peppermintTea):	& 0.60	& 0.4       	& 0.0433 \\
  wantsNotDrink(mary,proteinShake):	& 0.80	& 0.9831    	& 0.0182 \\
  wantsNotDrink(mary,redBull):	& 0.80	& 0.98      	& 0.0176 \\
  wantsNotDrink(mary,tea):	& 0.50	& 0.88      	& 0.0973 \\
  wantsNotDrink(mary,water):	& 0.50	& 0.1       	& 0.4047 \\
\midrule
                       & 	& 	& 2.3472 \\ 
\bottomrule
\end{tabularx}\medskip

We can see that not only the proposition that \textsc{Mary does not want to drink coffee} becomes much more manifest to Peter, but also the propositions to the effect that Mary wouldn't want similar tiredness-blocking drinks such as coke and Red Bull. Moreover, the manifestness of the assumptions that Mary is unlikely to drink fruit tea or peppermint tea are significantly decreased. In other words, we can see that Mary's utterance not only conveys a strong implicature about Mary not wanting to drink coffee, but also a variety of more weakely communicated implicatures.

In our model we treat this interpretive hypothesis as the speaker-intended one. We notice that in this interpretation, the presumed relevance expectations are satisfied in the sense that inferences changing the probability distribution of the queries of interest are available, and the knowledge base providing access to these inferences is not bloated, i.e. there are little or no superfluous assumptions in the knowledge graph.

\subsubsection{Interpretive hypothesis Int 2}
\label{sec:interpr-hypoth-int-2}

This interpretive hypothesis is chosen to illustrate what happens if we construct a hypothesis that does not satisfy the extant relevance expectations. To this end we cut the knowledge graph representing encyclopaedic knowledge to cover only assumptions about coffee, the only beverage mentioned directly in Peter's question.

\medskip\noindent
\begin{tabularx}{\linewidth}{lXXX}
\toprule
  Assumption (thought) & P (prior) & P (after) & KLD \\
\midrule
  wantsNotDrink(mary,coffee):	& 0.40	& 0.9208    	& 0.1873 \\
  wantsNotDrink(mary,coke):	& 0.50	& 0         	& inf \\
  wantsNotDrink(mary,fruitTea):	& 0.60	& 0         	& inf \\
  wantsNotDrink(mary,peppermintTea):	& 0.60	& 0         	& inf \\
  wantsNotDrink(mary,proteinShake):	& 0.80	& 0         	& inf \\
  wantsNotDrink(mary,redBull):	& 0.80	& 0         	& inf \\
  wantsNotDrink(mary,tea):	& 0.50	& 0         	& inf \\
  wantsNotDrink(mary,water):	& 0.50	& 0         	& inf \\\midrule
                       & 	& 	& inf \\
\bottomrule
\end{tabularx}\medskip

Notice that in this case, many queries of interest cannot be answered, leading to meaningless values of the KL divergence, which is set to infinity.

\subsubsection{Interpretive hypothesis Int 3}
\label{sec:interpr-hypoth-int-3}

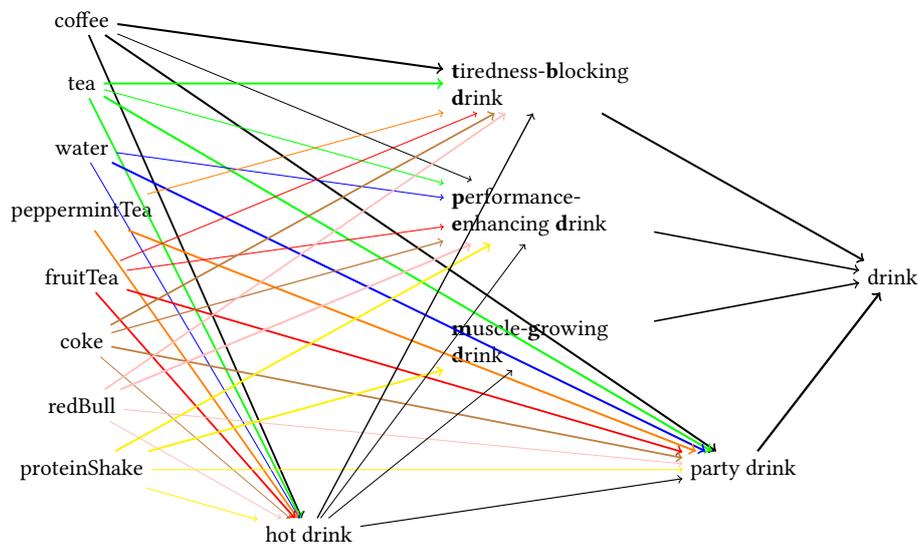
\begin{figure}
\resizebox{\textwidth}{!}{%
\begin{tikzpicture}
  \node (cof) at (0,4cm) {coffee};
  \node (tea) at (0,3cm) {tea};
  \node (wat) at (0,2cm) {water};
  \node (pep) at (0,1cm) {peppermintTea};
  \node (fru) at (0,0) {fruitTea};
  \node (cok) at (0,-1cm) {coke};
  \node (red) at (0,-2cm) {redBull};
  \node (pro) at (0,-3cm) {proteinShake};
  
  \node (tbd) at (7.2cm,3cm) {
    \begin{minipage}{3cm}
      \textbf{t}iredness-\textbf{b}locking \textbf{d}rink
    \end{minipage}};
  \node (ped) at (7.2cm,1cm) {
    \begin{minipage}{3cm}
      \textbf{p}erformance-\textbf{e}nhancing \textbf{d}rink
    \end{minipage}};
  \node (mgd) at (7.2cm,-1cm) {
    \begin{minipage}{3cm}
      \textbf{m}uscle-\textbf{g}rowing \textbf{d}rink
    \end{minipage}};
  
  \node (par) at (10.2cm,-3cm) {party drink};

  \node (hot) at (3.5cm,-4cm) {hot drink};
  
  \node (drk) at (12.5cm,0) {drink};

  \graph [edge quotes={fill=white, inner sep=1pt}] {
    (cof) ->[line width=0.9pt] (tbd) ;
    (cof) ->[line width=0.2pt] (ped) ;
    (cof) ->[line width=0.9pt] (par) ;
    (cof) ->[line width=0.8pt] (hot) ;

    (tea) ->[line width=0.85pt, green] (tbd) ;
    (tea) ->[line width=0.2, green] (ped) ;
    (tea) ->[line width=0.9, green] (par) ;
    (tea) ->[line width=0.8, green] (hot) ;

    (wat) ->[line width=0.9pt, blue] (par) ;
    (wat) ->[line width=0.1pt, blue] (ped) ;
    (wat) ->[line width=0.1pt, blue] (hot) ;
    
    (tbd) ->[line width=0.9pt] (drk) ;
    (ped) ->[line width=0.7pt] (drk) ;
    (mgd) ->[line width=0.7pt] (drk) ;

    (pep) ->[line width=0.4pt, orange] (tbd) ;
    (pep) ->[line width=0.9pt, orange] (par) ;
    (pep) ->[line width=0.8pt, orange] (hot) ;

    (fru) ->[line width=0.8pt, red] (par) ;
    (fru) ->[line width=0.1pt, red] (tbd) ;
    (fru) ->[line width=0.1pt, red] (ped) ;
    (fru) ->[line width=0.8pt, red] (hot) ;

    (cok) ->[line width=0.8pt, brown] (tbd) ;
    (cok) ->[line width=0.6pt, brown] (ped) ;
    (cok) ->[line width=0.8pt, brown] (par) ;
    (cok) ->[line width=0.1pt, brown] (hot) ;

    (red) ->[line width=0.9pt, pink] (ped) ;
    (red) ->[line width=0.8pt, pink] (tbd) ;
    (red) ->[line width=0.3pt, pink] (par) ;
    (red) ->[line width=0.1pt, pink] (hot) ;

    (pro) ->[line width=0.87pt, yellow] (mgd) ;
    (pro) ->[line width=0.87pt, yellow] (ped) ;
    (pro) ->[line width=0.1pt, yellow] (par) ;
    (pro) ->[line width=0.2pt, yellow] (hot) ;

    (par) ->[line width=1pt]  (drk) ;

    (hot) ->[line width=0.5pt] (par) ;
    (hot) ->[line width=0.6pt] (tbd) ;
    (hot) ->[line width=0.4pt] (ped) ;
    (hot) ->[line width=0.4pt] (mgd) ;

    };
\end{tikzpicture}}
\caption{Elaborate knowledge graph of encyclopaedic knowledge about energizing drinks and drinks offered at parties.}
\label{fig:kb-elaborated}
\end{figure}

This interpretive hypothesis involves a more elaborate knowledge graph of encyclopaedic knowledge about energizing drinks and drinks offered at parties in the sense that it adds a layer of classification into hot drinks (and, in consequence, cold drinks) along the path from individual beverages to the final category. (See Figure~\ref{fig:kb-elaborated}). 

However, this additional layer of encyclopaedic knowledge is not needed to satisfy relevance expectations. In other words, it is an interpretation that is more costly to construct than Int 1.

\medskip\noindent
\begin{tabularx}{\linewidth}{lXXX}
\toprule
  Assumption (thought) & P (prior) & P (after) & KLD \\
\midrule
  wantsNotDrink(mary,coffee):	& 0.40	& 0.97503616	& 0.2186 \\
  wantsNotDrink(mary,coke):	& 0.50	& 0.926848  	& 0.1183 \\
  wantsNotDrink(mary,fruitTea):	& 0.60	& 0.69107248	& 0.0063 \\
  wantsNotDrink(mary,peppermintTea):	& 0.60	& 0.81088   	& 0.0302 \\
  wantsNotDrink(mary,proteinShake):	& 0.80	& 0.98599328	& 0.0188 \\
  wantsNotDrink(mary,redBull):	& 0.80	& 0.981712  	& 0.0180 \\
  wantsNotDrink(mary,tea):	& 0.50	& 0.962176  	& 0.1349 \\
  wantsNotDrink(mary,water):	& 0.50	& 0.17704   	& 0.1962 \\
\midrule
  	& 	& 	& 0.7411 \\
\bottomrule
\end{tabularx}\medskip

In the results we can see that the stronger and weaker implicatures can be derived in similar ways to Int 1. Yet the utility of the utterance as indicated by the KL divergence between $P(\mathrm{prior})$ and $P(\mathrm{after})$ is lower in Int 3 than in Int 1 due to the additional probability calculations about the additional edges in the graph.

In classical relevance theory, Int 3 is said to be less relevant than Int 1 because the processing effort involved exceeds that involved in processing Int 1, and this effort is not offset by additional cognitive effects contributing to the satisfaction of relevance expectations. In the probabilistic computational model we are developing, this difference is captured by the lower value of the sum total of the KL divergence. However, more costly to construct interpretive hypotheses do not always affect the KL measure, as we shall see directly when we turn to discuss yet another interpretive hypothesis.

\subsubsection{Interpretive hypothesis Int 4}
\label{sec:interpr-hypoth-int-4}

Int 4 includes the same encyclopaedic knowledge graph about beverages as in Int 1. In addition to this it provides another graph that allows to draw conclusions about whether Mary, if she does not want to drink coffee after 6PM, may be a particular health-conscious person. This second graph does not affect the probability calculations involving the edges of the graph describing the encyclopaedic knowledge about beverages.

However, if we assume that Peter's relevance expectations are the same as in Int 1, then the additional possible inferences do not affect the answers available to the queries corresponding the relevance expectations:

\medskip\noindent
\begin{tabularx}{\linewidth}{lXXX}
\toprule
  Assumption (thought)               & P (prior)     & P (after)  & KLD \\
\midrule
  wantsNotDrink(mary,coffee):	& 0.40	& 0.9208    	& 0.1873 \\
  wantsNotDrink(mary,coke):	& 0.50	& 0.92      	& 0.1151 \\
  wantsNotDrink(mary,fruitTea):	& 0.60	& 0.0199    	& 1.4636 \\
  wantsNotDrink(mary,peppermintTea):	& 0.60	& 0.4       	& 0.0433 \\
  wantsNotDrink(mary,proteinShake):	& 0.80	& 0.9831    	& 0.0182 \\
  wantsNotDrink(mary,redBull):	& 0.80	& 0.98      	& 0.0176 \\
  wantsNotDrink(mary,tea):	& 0.50	& 0.88      	& 0.0973 \\
  wantsNotDrink(mary,water):	& 0.50	& 0.1       	& 0.4047 \\
\midrule
                                     & 	& 	& 2.3472 \\
\bottomrule
\end{tabularx}\medskip

The KL measure of this interpretation Int 4 is the same as that of Int 1. Int 4 and Int 1 differ in another dimension: Int 4 has more so-called \emph{atoms} in the ProbLog program. In  logic programming in languages such as Prolog and ProbLog, an \emph{atom} is essentially a predicate representing a proposition that is assumed to hold true (with a certain probability) in the model of the world that the program is representing. The set of such atoms in a ProbLog program is called its \emph{Herbrand base} \parencite[Section 2.1]{flach1994simply}. Since Int 4 has an additional encyclopaedic knowledge graph over against Int 1, its Herbrand base is larger. It is natural to assume that  a larger Herbrand base means increased processing effort for processing this hypothesis.

Thus we have another way to differentiate two interpretive hypotheses according to the processing effort they incur: when the respective hypotheses do not differ in their KL measure, they will differ in the size of their Herbrand base.

\subsubsection{Interpretive hypothesis Int 5}
\label{sec:interpr-hypoth-int-5}

Int 5 contains the same knowledge base as Int 4. This time, we assume that information about Mary's being a particular health-conscious person is relevant to Peter if available (perhaps for drawing further conclusions not just about beverages Mary may or may not accept but also about what sort of foods or snacks Mary may appreciate at the party). Thus, the additional knowledge graph contributes additional possible inferences satisfying Peter's relevance expectations. Consequently, this interpretive hypothesis is more relevant to Peter than Int 1, as the increased processing effort involved in utilising the additional contextual knowledge satisfies more relevance expectations. This is reflected in a greater KL measure than the one in Int 1.

\medskip\noindent
\begin{tabularx}{\linewidth}{lllX}
\toprule
  Assumption (thought)               & P (prior)     & P (after)  & KLD \\
\midrule
  healthconscious(mary):	& 0.50	& 0.82228   	& 0.0735 \\
  wantsNotDrink(mary,coffee):	& 0.40	& 0.9208    	& 0.1873 \\
  wantsNotDrink(mary,coke):	& 0.50	& 0.92      	& 0.1151 \\
  wantsNotDrink(mary,fruitTea):	& 0.60	& 0.0199    	& 1.4636 \\
  wantsNotDrink(mary,peppermintTea):	& 0.60	& 0.4       	& 0.0432 \\
  wantsNotDrink(mary,proteinShake):	& 0.80	& 0.9831    	& 0.0182 \\
  wantsNotDrink(mary,redBull):	& 0.80	& 0.98      	& 0.0176 \\
  wantsNotDrink(mary,tea):	& 0.50	& 0.88      	& 0.0973 \\
  wantsNotDrink(mary,water):	& 0.50	& 0.1       	& 0.4047 \\
\midrule
                                     & 	& 	& 2.4207 \\
\bottomrule
\end{tabularx}\medskip

\subsubsection{Comprehension modelling as optimization model}
\label{sec:compr-modell-as}

The preceding discussion shows that we can conceptualize the comprehension of implicatures in the form of an optimization model having the following structure:

\begin{enumerate}
  \item Eliminate interpretive hypotheses from consideration that do not satisfy all relevance expectations (such as Int 2).

  \item If interpretive hypotheses have the same KL measure, consider only the one with the smallest Herbrand base.

  \item From among the remaining interpretive hypotheses, accept the one with the highest KL measure.
\end{enumerate}

This optimization model\footnote{`\emph{as-if}-optimization' in the terms of \textcite[64]{gigerenzerFastFrugalHeuristics2004a}
} allows us to construct quantitative predictions about the interpretation that the relevance-theoretic comprehension heuristic would pick out. These quantitative predictions could be used in experiments investigating the heuristic nature of relevance-guided comprehension.

\section{Conclusion}
\label{sec:conclusion}

In this paper we have developed an \emph{as-if-}optimization model of implicature comprehension that captures the insights of relevance theory in probabilistic terms and implemented a computational model of it in the probabilistic programming language ProbLog. This model can serve as a tool to investigate the heuristic nature of the relevance-theoretic comprehension procedure by way of comparing experimental results with quantitative model predictions.

The model shares with classical relevance theory its focus on the interpretive hypothesis-evaluation part of non-demonstrative inference processes, assuming that the hypothesis-generation part of non-demonstrative inference is inamenable to explanatory study \textcite[68]{sperberRelevance1995}. In consequence, we require candidate interpretive hypotheses of utterances to be manually constructed by the experimenters/programmers.

Moreover, this model focusses on the audience's part in communication, the comprehension of utterances, as does classical relevance theory. The reason for this is that given the relevance-seeking nature of human minds, speakers must take this nature into account if they want to be understood and choose their utterances such that the audience only needs to follow where their relevance expectation-satisfaction search will lead them. This means that responsibilities, and sophistication in cognitive processes, are unevenly distributed, making the comprehension phase of utterance interpretation more amenable to study. This does not mean, however, that the speaker's perspective is not important to study; on the contrary, extending a relevance-theoretic account of communication to the speaker's perspective is a strongly desirable goal. For this goal to be reached, it is important to gain a better understanding of the inference processes involved in metacognition: how to cognizers (and in particular, communicators) estimate what expatations of relevance others have?

Other areas where our model needs extension include the modelling of higher-level explicature comprehension (speech act and propositional attitude recognition, modal meanings) as well as to instances of metarepresentational utterance use, as for example in irony, direct and indirect quotation, free indirect speech and thought reports. This will be left to future research.

\printbibliography[heading=bibintoc]

\appendix

\section{Grice on Implicatures}
\label{sec:h.-paul-grice}

The philosopher Paul Grice fameously introduced the term \emph{Implicature} in his William James Lectures of 1967 \parencite{griceWilliamJamesLectures1967}, published in its most recent form in the chapter \emph{Logic and Conversation} in \textcite{griceStudiesWayWords1989}.

\pex<griceimplicaturefirst>
  Suppose that A and B are talking about a mutual friend, C, who is now working in a bank. A asks B how C is getting on in his job, and B replies, \emph{Oh quite well, I think; he likes his colleagues, and he hasn't been to prison yet.} \parencite[24]{griceStudiesWayWords1989}
\xe

Grice goes on to say that B can be informally said to have implied, suggested or meant ``\ldots{}any one of such things as that C is the sort of person likely to yield to the temptation provided by his occupation, that C's colleagues are really very unpleasant and treacherous people, and so forth.''\parencite[24]{griceStudiesWayWords1989} In order to enhance clarity of the discussion, he introduces the technical term \emph{implicate}: B implicates -- rather than says -- that C is likely to succumb to the temptation of dishonesty, the proposition \textsc{C is likely to succumb to the temptation of dishonesty} is an \emph{implicature} of (or rather, one of the implicatures of) B's utterance \emph{he hasn't been to prison yet} in the example. \emph{Saying} is in essence expressing the meaning of the conventional meaning of the sentence. So we can say that in the above example, B \emph{says} that \emph{C hasn't been to prison yet} and \emph{implicates} that \emph{C is likely to succumb to the temptation of dishonesty}, or that B's utterance conveys the meaning that \emph{C is likely to succumb to the temptation of dishonesty} by way of \emph{implicature.}

In other words, implicatures (more precisely \emph{conversational implicatures} in Grice's terminology) are aspects of the communicator's intended meaning which can not be read off the conventional, linguistically encoded sentence meaning that the communicator has committed to conveying.\footnote{%
  This reference to speaker's commitment is important because Grice claims that there are some implicatures which are connected to the conventional meaning of words such as \emph{therefore}, which he calls \emph{conventional implicatures.} However, these conventional implicatures are cancellable and do not contribute to the truth conditions of the utterance. }
Implicatures arise in the use of language in  conversation, in communication. But how can audiences reliably comprehend such implicated meanings? Grice's answer was that conversation (communication) between rational agents is a cooperative endeavour, and what this amounts to is spelled out by a set of maxims of Quantity, Quality, Relation and Manner:

\pex<gricemaxims> The Conversational Maxims of \textcite[26--27]{griceStudiesWayWords1989}: 
\begin{itemize} 
\item[] \emph{Cooperative Principle: } Make your contribution such as is required,
at the stage at which it occurs,
by the acepted purpose or direction of the talk exchange
in which you are engaged. 
\item[] \emph{Quantity: }
\begin{enumerate}
\item Make your contribution as informative as is required
  (for the current purposes of the eschange)
\item Do not make your contribution more informative than is required.
\end{enumerate} 
\item[] \emph{Quality: } Try to make your contribution one that is true. 
\begin{enumerate}
\item Do not say what you believe to be false.
\item Do not say that for which you lack adequate evidence.
\end{enumerate}
\item[] \emph{Relation: } Be relevant. 
\item[] \emph{Manner: }
  \begin{enumerate}
  \item Avoid obscurity of expression.
  \item Avoid ambiguity.
  \item Be brief (avoid unnecessary prolixity).
  \item Be orderly. 
  \end{enumerate}

\end{itemize}
\xe

A communicator conveying an implicature necessarily produces an utterance whose conventional (literal) meaning does not adhere to these maxims, at least not to all of them. However, the communicator can count on the audience's ability to see what the utterance might imply in the context at hand; if this implied import in turn satisfies the maxims, then the audience is justified to assume that the communicator intended to implicate this implied meaning, i.e. to convey this meaning as an implicature. Grice explains the pattern of inference for the comprehension of implicatures as follows:

\pex<griceimplicaturedef> 
  A man who, by (in, when) saying (or making as if to say) that \emph{p} has implicated that \emph{q}, may be said to have conversationally implicated that \emph{q}, provided that (1) he is to be presumed to be observing the conversational maxims, or at least the Cooperative Principle; (2) the supposition that he is aware that, or thinks that, \emph{q} is required in order to make his saying or making as if to say \emph{p} (or doing so in \emph{those} terms) consistent with this presumption; and (3) the speaker thinks (and would expect the hearer to think that the speaker thinks) that it is within the competence of the hearer to work out, or grasp intuitively, that the supposition mentioned in (2) is required. \parencite[31]{griceStudiesWayWords1989}
\xe

Applying this pattern to his introductory example (\getfullref{griceimplicaturefirst}), Grice writes the following:

\pex<griceimplicatureexample> In a suitable setting A might reason as follows:
``(1) B has apparently violated the maxim e relevant'  and so may be regarded as having flouted one of the maxims conjoining perspicuity, yet I have no reason to suppose that he is option out from the operation of the Cooperative Principle; (2) given the circumstances, I can regard his irrelevance as only apparent if, and only if, I suppose him to think that C is potentially dishonest; (3) B knows that I am capable of working out step (2). So B implicates that C is potentially dishonest.'' \parencite[31]{griceStudiesWayWords1989}
\xe

There are two points to notice: first, it remains unexplained in (\getfullref{griceimplicaturedef}) how the audience arrives at \emph{q} given \emph{p}. The reasoning process would certainly need to include some additional premises not derivable from \emph{p}, and the question of how to choose these premises would need to be addressed \parencite[378]{wilsonInferenceImplicature1991}. Second, notice that although this inference pattern is presented as a schema for a fully conscious and elaborate reasoning process, Grice does not claim that audiences necessarily apply this reasoning processes. In fact, he grants that audiences can often intuit implicatures subconsciously. However, he insists that hearers must be able to work out the implicatures according to this schema. In other words, this working out schema for implicatures is not intended to provide a psychologically plausible inference pattern or procedure to identify implicatures. Rather, it serves as an analytical tool for theorists to describe the phenomenon.

\section{Code listing: Optimally relevant interpretive hypothesis in ProbLog -- Int 1, Int 2}
\label{sec:code-list-optim}

Interpretive hypothesis Int 1 is implemented as a ProbLog program as listed below.
Int 2 differs from Int 1 by ommitting lines 11--29. 

\begin{lstlisting}

% Interpretive hypothesis 1: satisfying relevance expectations, optimal

% Modelling a web of easily accessible encyclopaedic assumptsions
% as probabilistic graphs from beverages via categorizing nodes
% (e.g. tirednessBlockingDrink, partyDrink) 
% to the superclass drink.
0.9::edge(coffee, tirednessBlockingDrink).
0.2::edge(coffee, performanceEnhancingDrink).
0.8::edge(coffee, partyDrink).
0.9::edge(water, partyDrink).
0.1::edge(water, performanceEnhancingDrink).
0.85::edge(tea, tirednessBlockingDrink).
0.2::edge(tea, performanceEnhancingDrink).
0.9::edge(tea, partyDrink).
0.4::edge(peppermintTea, tirednessBlockingDrink).
0.9::edge(peppermintTea, partyDrink).
0.8::edge(fruitTea, partyDrink).
0.01::edge(fruitTea, tirednessBlockingDrink).
0.01::edge(fruitTea, performanceEnhancingDrink).
0.8::edge(coke, tirednessBlockingDrink).
0.6::edge(coke, performanceEnhancingDrink).
0.8::edge(coke, partyDrink).
0.9::edge(redBull, performanceEnhancingDrink).
0.8::edge(redBull, tirednessBlockingDrink).
0.3::edge(redBull, partyDrink).
0.87::edge(proteinShake, muscleGrowingDrink).
0.87::edge(proteinShake, performanceEnhancingDrink).
0.1::edge(proteinShake, partyDrink).
edge(partyDrink, beverage).


% Definition of person
person(X) :- path(X, person).
% Mary is a person
edge(mary, person).

% The word energizingDrink could convey the concepts
% tirednessBlockingDrink; performanceEnhancingDrink; muscleGrowingDrink.
energizingDrink(X) :- 0.9::path(X,tirednessBlockingDrink);
                        0.7::path(X,performanceEnhancingDrink);
                          0.7::path(X,muscleGrowingDrink).

% For categorization, we need performanceEnhancingDrink/1, etc.
performanceEnhancingDrink(X) :- path(X,performanceEnhancingDrink).

muscleGrowingDrink(X) :- path(X,muscleGrowingDrink).


% Defining paths and leafs (outer nodes) for inference
path(X,Y) :- edge(X,Y).
path(X,Y) :- edge(X,Z),
             Y \= Z,
             path(Z,Y).

% Description of the communicator's ostensive act (utterance)
say(P,S) :- person(P), sentence(S).

% Proposition which is the result of the developlment of the logical form of the utterance = explicature;
% masking a lot of details
% The situation that P does not want to drink X is asserted to be true in the world
% if P utters the sentence "I don't drink energizing drinks..."
% and X can be said to be of the category "energizing drink". 
wantsNotDrink(P,X) :- say(P,sentence(idontDrinkED)), energizingDrink(X).
wantsNotDrink(P,X) :- say(P,sentence(idontDrinkPeD)), performanceEnhancingDrink(X).
wantsNotDrink(P,X) :- say(P,sentence(idontDrinkMgD)), muscleGrowingDrink(X).
wantsNotDrink(P,X) :- say(P,sentence(idontDrinkCoffee)), X=coffee.


% Some sentences Mary could have said. Need some prior probability less than 1,
% otherwise we get a message about inconsistent evidence. 
0.01::sentence(idontDrinkED).
0.01::sentence(idontDrinkPeD).
0.01::sentence(idontDrinkMgD).
0.01::sentence(idontDrinkCoffee).

% Sentences Mary may have said.
% Since we want to use a description of what Mary said as evidence,
% we need to assign a certain prior probability to these descriptions,
% otherwise we get an error message about inconsistent evidence.
0.01::say(mary,sentence(idontDrinkED)).
0.01::say(mary,sentence(idontDrinkPeD)).
0.01::say(mary,sentence(idontDrinkMgD)).
0.01::say(mary,sentence(idontDrinkCoffee)).
  
% Mary provides Peter with partial evidence for her informative intention
% by uttering the sentence modeled above. Choose one, or none.
evidence(say(mary,sentence(idontDrinkED))).
% evidence(say(mary,sentence(idontDrinkPeD))).
% evidence(say(mary,sentence(idontDrinkMgD))).
% evidence(say(mary,sentence(idontDrinkCoffee))).

% Inference means querying about the beverages 
query(wantsNotDrink(mary,coffee)).
query(wantsNotDrink(mary,tea)).
query(wantsNotDrink(mary,peppermintTea)).
query(wantsNotDrink(mary,fruitTea)).
query(wantsNotDrink(mary,coke)).
query(wantsNotDrink(mary,redBull)).
query(wantsNotDrink(mary,water)).
query(wantsNotDrink(mary,proteinShake)).

\end{lstlisting}

\section{Code listing: Int 3}
\label{sec:int-3}

As in Int 1 until line 31.
The following probabilistic edges are added at this point:

\begin{lstlisting}[firstnumber=32]
% Hot vs cold drink classification; irrelevant to expectations
0.8::edge(coffee, hotDrink).
0.1::edge(water, hotDrink).
0.8::edge(tea, hotDrink).
0.8::edge(peppermintTea, hotDrink).
0.8::edge(fruitTea, hotDrink).
0.1::edge(coke, hotDrink).
0.1::edge(redBull, hotDrink).
0.2::edge(proteinShake, hotDrink).

0.5::edge(hotDrink, partyDrink).
0.6::edge(hotDrink, tirednessBlockingDrink).
0.4::edge(hotDrink, performanceEnhancingDrink).
0.4::edge(hotDrink, muscleGrowingDrink).

\end{lstlisting}

\section{Code listing: Int 4 and Int 5}
\label{sec:int-4-int}

Int 4 and Int 5 differ from one another only in the extant relevance expectations: 
Int 5 adds the query \verb+query(healthconscious(mary))+.
Other than that Int 4 and Int 5 are identical in their knowledge graphs.
The implementation equals that of Int 1 up to line 31.
At this point, the following second knowledge graph is added:

\begin{lstlisting}[firstnumber=32]
% Additional encyclopaedic web for potential additional inference
% Whether Mary is healthconscious may be relevant to Peter,
% it could also affect his next move. 
0.5::edge(person(mary), healthconscious).
0.2::edge(notDrinkCoffeeBefore6PM, healthconscious).
0.7::edge(notDrinkCoffeeAfter6PM, healthconscious).

edge(person(mary), notDrinkCoffeeAfter6PM) :- wantsNotDrink(mary, coffee).

\end{lstlisting}

\end{document}